\documentclass{article}

\usepackage{arxiv}
\usepackage[T1]{fontenc}
\usepackage{pdfpages}
\renewcommand{\shorttitle}{Physical regime changes expose formula misuse in LLMs}
\usepackage{setspace}

\usepackage[numbers,sort&compress]{natbib}

\usepackage{amsmath,amssymb,amsfonts}
\usepackage{siunitx}

\usepackage{graphicx}
\graphicspath{{figures/}}

\usepackage{booktabs}
\usepackage{array}
\usepackage{multirow}

\usepackage{url}
\usepackage[hidelinks]{hyperref}
\usepackage{xcolor}
\usepackage[utf8]{inputenc}
\usepackage{float}

\usepackage{caption}
\usepackage{etoolbox}
\DeclareCaptionFont{figurecaption}{\fontsize{10}{12}\selectfont}
\AtBeginEnvironment{table}{\singlespacing\fontsize{9}{11}\selectfont
  \setlength{\tabcolsep}{4pt}\renewcommand{\arraystretch}{1.12}}
\AtBeginEnvironment{tabular}{\fontsize{8.5}{10.5}\selectfont}
\AtBeginEnvironment{tabularx}{\fontsize{9}{11}\selectfont}
\AtBeginEnvironment{longtable}{\fontsize{8}{10}\selectfont}
\AtBeginEnvironment{quote}{\raggedright\fontsize{9}{11}\selectfont}
\AtEndEnvironment{table}{\par\vspace{5pt}}
\newcommand{\nLocalConfigs}{16} %
\newcommand{\nTemplatesDiag}{69} %
\newcommand{\nClustersDiag}{31} %
\newcommand{\rewordMaxChange}{4.7} %
\newcommand{\dFiveGap}{12.7} %
\newcommand{\dFiveGapLo}{4.4} %
\newcommand{\dFiveGapHi}{21.3} %
\newcommand{\rArmHoursQwenMin}{2} %
\newcommand{\rArmHoursQwenMax}{3} %
\newcommand{\rArmHoursQwenBMin}{6} %
\newcommand{\rArmHoursQwenBMax}{8} %
\newcommand{\sThreeN}{16} %
\newcommand{\sThreeMet}{6} %
\newcommand{\judgeSaidNoQwen}{1.6} %
\newcommand{\judgeSaidNoQwenB}{18.1} %
\newcommand{\judgeTCQwenB}{-6.0} %
\newcommand{\judgeTCQwenBLo}{-26.9} %
\newcommand{\judgeTCQwenBHi}{14.7} %
\newcommand{\judgeTCDS}{2.5} %
\newcommand{\judgeTCDSLo}{-1.9} %
\newcommand{\judgeTCDSHi}{8.4} %
\newcommand{\costRegimeDefined}{97} %
\newcommand{\costRegimeMedian}{51} %
\newcommand{\costRegimeMedianLo}{45} %
\newcommand{\costRegimeMedianHi}{62} %
\newcommand{\costRegimeShare}{52} %
\newcommand{\computeRuns}{68} %
\newcommand{\computeGPUh}{5,752} %
\newcommand{\insideMinusOutsideMin}{13} %
\newcommand{\insideMinusOutsideMax}{57} %
\newcommand{\givenMinusInsideMin}{-8} %
\newcommand{\givenMinusInsideMax}{14} %
\newcommand{\flagMinusOutsideMin}{0.7} %
\newcommand{\flagMinusOutsideMax}{8.4} %
\newcommand{\frontierInsideMinusOutside}{27.6} %
\newcommand{\frontierInsideMinusOutsideLo}{15.5} %
\newcommand{\frontierInsideMinusOutsideHi}{40.3} %
\newcommand{\shortcutShareMin}{32} %
\newcommand{\shortcutShareMax}{69} %
\newcommand{\frontierShortcutShare}{46} %
\newcommand{\nTrapsCost}{328} %
\newcommand{\nTrapsCostDefined}{280} %
\newcommand{\costMedianDeviation}{55} %
\newcommand{\costMedianDeviationLo}{50} %
\newcommand{\costMedianDeviationHi}{65} %
\newcommand{\costShareGeTwo}{52} %
\newcommand{\costCountGtTen}{19} %
\newcommand{\lineBroadeningFactor}{29} %
\newcommand{\teacherGate}{6} %
\newcommand{\teacherProbeZ}{14} %
\newcommand{\teacherKnownLarge}{0} %
\newcommand{\teacherKnownMoe}{1} %
\newcommand{\nUniversalFail}{7} %
\newcommand{\coTwoDoublingLinear}{5.35} %
\newcommand{\coTwoDoublingLog}{3.71} %
\newcommand{\sunPhotometerDeviation}{21} %
\newcommand{\atlasTemplateShare}{44} %
\newcommand{\atlasPermP}{0.0001} %
\newcommand{\judgedNoShortcut}{53} %
\newcommand{\judgedYesShortcut}{52} %
\newcommand{\recallQwen}{1.4} %
\newcommand{\recallQwenLo}{-5.2} %
\newcommand{\recallQwenHi}{8.1} %
\newcommand{\handbookNone}{23} %
\newcommand{\handbookWith}{54} %
\newcommand{\handbookGiven}{97} %
\newcommand{\detectH}{32} %
\newcommand{\detectF}{6} %
\newcommand{\detectHminusF}{26.5} %
\newcommand{\detectHminusFLo}{16.5} %
\newcommand{\detectHminusFHi}{37.4} %
\newcommand{\rlGainTestId}{23.8} %
\newcommand{\rlGainTestBook}{24.1} %
\newcommand{\rlGainCrossDomain}{19.3} %
\newcommand{\rlGainScaffoldStripped}{22.6} %
\newcommand{\randomRewardNineMin}{-3.9} %
\newcommand{\randomRewardNineMax}{3.0} %
\newcommand{\nProbeRuns}{36} %
\newcommand{\nCrossDomain}{131} %
\newcommand{\heldGainQwenMin}{2} %
\newcommand{\heldGainQwenMax}{9} %
\newcommand{\asymHeldQwenA}{-18.7} %
\newcommand{\asymHeldQwenALo}{-29.3} %
\newcommand{\asymHeldQwenAHi}{-8.4} %
\newcommand{\trapsBaseQwen}{2.2} %
\newcommand{\trapsAQwen}{16.8} %
\newcommand{\trapsAQwenLo}{9.2} %
\newcommand{\trapsAQwenHi}{25.0} %
\newcommand{\trapsBaseQwenB}{34.8} %
\newcommand{\trapsAQwenB}{54.9} %
\newcommand{\trapsAQwenBLo}{43.5} %
\newcommand{\trapsAQwenBHi}{66.3} %
\newcommand{\transferArmsMetQwenB}{3} %
\newcommand{\pTwoFminusAQwenBSeed}{54.9} %
\newcommand{\pTwoFminusAQwenBSeedLo}{44.3} %
\newcommand{\pTwoFminusAQwenBSeedHi}{65.4} %
\newcommand{\frontierN}{46} %
\newcommand{\frontierAQwenB}{54.9} %
\newcommand{\frontierDSOff}{56.5} %
\newcommand{\frontierDiffDS}{-1.6} %
\newcommand{\frontierDiffDSLo}{-12.7} %
\newcommand{\frontierDiffDSHi}{9.2} %
\newcommand{\frontierGPTFiveFiveOff}{78.3} %
\newcommand{\frontierGPTFiveFourOff}{71.0} %
\newcommand{\frontierMiniMaxOff}{14.5} %
\newcommand{\frontierOnMin}{97} %
\newcommand{\frontierOnMax}{99} %
\newcommand{\frontierTemplateGapMin}{12} %
\newcommand{\frontierTemplateGapMax}{32} %
\newcommand{\sharedN}{38} %
\newcommand{\gapClosedLarge}{62} %
\newcommand{\indepRegimeN}{54} %
\newcommand{\indepGeometryN}{56} %
\newcommand{\indepParentN}{97} %
\newcommand{\indepRegimeGain}{20.4} %
\newcommand{\indepRegimeGainLo}{7.4} %
\newcommand{\indepRegimeGainHi}{33.3} %
\newcommand{\nZeroHit}{14} %
\newcommand{\nZeroHitNotLearnedAny}{13} %
\newcommand{\nZeroHitQwenB}{5} %
\newcommand{\nZeroHitLearnedQwenB}{2} %
\newcommand{\nZeroHitPooled}{19} %
\newcommand{\nZeroHitLearnedPooled}{3} %
\newcommand{\recallQwenB}{11.3} %
\newcommand{\recallQwenBLo}{4.9} %
\newcommand{\recallQwenBHi}{18.4} %
\newcommand{\pOneAUC}{0.91} %
\newcommand{\pOneAUCLo}{0.84} %
\newcommand{\pOneAUCHi}{0.97} %
\newcommand{\pOneFisherP}{1.8 \times 10^{-5}} %
\newcommand{\pcurveQwenBinZeroLearned}{6} %
\newcommand{\pcurveQwenBinZeroN}{32} %
\newcommand{\pcurveQwenBinThreeLearned}{10} %
\newcommand{\pcurveQwenBinThreeN}{10} %
\newcommand{\shortcutHeldA}{20.4} %
\newcommand{\shortcutHeldALo}{11.3} %
\newcommand{\shortcutHeldAHi}{30.3} %
\newcommand{\shortcutHeldR}{-4.4} %
\newcommand{\shortcutHeldRLo}{-11.1} %
\newcommand{\shortcutHeldRHi}{1.7} %
\newcommand{\shortcutBaseLevel}{31} %
\newcommand{\shortcutLevelbase}{29.8} %
\newcommand{\shortcutLevelA}{74.5} %
\newcommand{\shortcutLevelP}{16.1} %
\newcommand{\doseHOne}{61.3} %
\newcommand{\doseHOneLo}{38.5} %
\newcommand{\doseHOneHi}{82.9} %
\newcommand{\doseHOneRounded}{61} %
\newcommand{\doseMechanisms}{17} %
\newcommand{\doseQwenDosedLearned}{16} %
\newcommand{\doseRLminusWarm}{39.4} %
\newcommand{\doseRLminusWarmLo}{21.4} %
\newcommand{\doseRLminusWarmHi}{59.0} %
\newcommand{\doseWarmOnlyLearned}{5} %
\newcommand{\doseQwenDosedN}{26} %
\newcommand{\doseQwenZeroLearned}{0} %
\newcommand{\doseQwenZeroN}{8} %
\newcommand{\doseOneLearned}{5} %
\newcommand{\doseOneN}{9} %
\newcommand{\pooledHighPLearned}{22} %
\newcommand{\pooledHighPN}{22} %
\newcommand{\doseRandP}{0.002} %
\newcommand{\doseQwenDOneWarm}{3.3} %
\newcommand{\doseQwenDOneRL}{55.0} %
\newcommand{\doseQwenDFourWarm}{50.6} %
\newcommand{\doseQwenDFourRL}{87.5} %
\newcommand{\doseMinDFourN}{7} %
\newcommand{\doseMinDosedLearned}{83} %
\newcommand{\doseMinHOne}{50.0} %
\newcommand{\doseMinHOneLo}{-20.0} %
\newcommand{\doseMinHOneHi}{94.4} %
\newcommand{\doseTextHOne}{6.5} %
\newcommand{\doseTextHOneLo}{-9.0} %
\newcommand{\doseTextHOneHi}{20.0} %
\newcommand{\opdP}{70.7} %
\newcommand{\opdPLo}{40.0} %
\newcommand{\opdPHi}{95.4} %
\newcommand{\opdPRounded}{71} %
\newcommand{\opdS}{0.0} %
\newcommand{\opdPshuf}{0.0} %
\newcommand{\opdPminusPshufAll}{35.2} %
\newcommand{\opdPminusPshufAllLo}{26.6} %
\newcommand{\opdPminusPshufAllHi}{44.1} %
\newcommand{\opdPminusPshufMain}{70.7} %
\newcommand{\opdPminusPshufMainLo}{40.7} %
\newcommand{\opdPminusPshufMainHi}{96.4} %
\newcommand{\opdPminusS}{70.7} %
\newcommand{\opdPminusSLo}{40.4} %
\newcommand{\opdPminusSHi}{95.4} %
\newcommand{\opdMainSet}{7} %
\newcommand{\sTwoUnits}{65} %
\newcommand{\samplesPerMechanism}{1280} %
\newcommand{\nRegimeTraps}{46} %
\newcommand{\aucStudentA}{0.89} %
\newcommand{\aucTeacherA}{0.51} %
\newcommand{\aucDiffA}{0.38} %
\newcommand{\aucDiffALo}{0.22} %
\newcommand{\aucDiffAHi}{0.53} %
\newcommand{\aucStudentP}{0.62} %
\newcommand{\aucTeacherP}{0.86} %
\newcommand{\aucDiffP}{-0.24} %
\newcommand{\aucDiffPLo}{-0.41} %
\newcommand{\aucDiffPHi}{-0.07} %
\newcommand{\rFourLearnedCells}{166} %
\newcommand{\rFourMaxInsideDrop}{10} %
\newcommand{\rFourInsideStartQwenA}{37} %
\newcommand{\rFourOutsideStartQwenA}{8} %
\newcommand{\rFourInsideStartQwenBA}{76} %
\newcommand{\rFourOutsideStartQwenBA}{15} %
\newcommand{\rFourInsideStartQwenDose}{49} %
\newcommand{\rFourInsideEndQwenDose}{100} %
\newcommand{\rFourOutsideStartQwenDose}{14} %
\newcommand{\rFourInsideStartP}{50} %
\newcommand{\rFourInsideEndP}{70} %
\newcommand{\rFourOutsideStartP}{7} %
\newcommand{\rFourSwapCells}{0} %
\newcommand{\rFourDinsideQwenA}{58.0} %
\newcommand{\rFourDinsideQwenALo}{46.1} %
\newcommand{\rFourDinsideQwenAHi}{69.2} %
\newcommand{\rFourDhfQwenA}{-9.5} %
\newcommand{\rFourDhfQwenALo}{-19.5} %
\newcommand{\rFourDhfQwenAHi}{-0.3} %
\newcommand{\rFourDhfQwenBA}{-15.2} %
\newcommand{\rFourDhfQwenBALo}{-27.3} %
\newcommand{\rFourDhfQwenBAHi}{-4.2} %
\newcommand{\rFourDoseTraps}{19.0} %
\newcommand{\rFourDoseMinusWA}{3.8} %
\newcommand{\rFourDoseMinusWALo}{-2.2} %
\newcommand{\rFourDoseMinusWAHi}{9.8} %
\newcommand{\rFourWATraps}{15.2} %
\newcommand{\rhoMinQwen}{0.78} %
\newcommand{\rhoMinQwenLo}{0.64} %
\newcommand{\rhoMinQwenHi}{0.87} %
\newcommand{\rhoQwenQwenB}{0.47} %
\newcommand{\rhoQwenQwenBLo}{0.25} %
\newcommand{\rhoQwenQwenBHi}{0.65} %
\newcommand{\zeroInAllFamilies}{3} %
\newcommand{\warmupGain}{33.1} %
\newcommand{\warmupGainLo}{20.7} %
\newcommand{\warmupGainHi}{46.4} %
\newcommand{\counterfactualGain}{3.1} %
\newcommand{\counterfactualGainLo}{-0.5} %
\newcommand{\counterfactualGainHi}{7.7} %
\newcommand{\pTwoFminusAQwen}{25.3} %
\newcommand{\pTwoFminusAQwenLo}{15.8} %
\newcommand{\pTwoFminusAQwenHi}{35.6} %
\newcommand{\extSteps}{200} %
\newcommand{\extMechs}{eight} %
\newcommand{\extZeroPassStart}{six} %
\newcommand{\extZeroPassEnd}{seven} %
\newcommand{\extNoSignal}{seven} %
\newcommand{\luckFunctionalRare}{50} %
\newcommand{\luckFisherP}{2.8 \times 10^{-9}} %
\newcommand{\rhoQwenBLarge}{0.76} %
\newcommand{\rhoQwenBLargeLo}{0.55} %
\newcommand{\rhoQwenBLargeHi}{0.90} %
\newcommand{\nZeroHitLarge}{5} %
\newcommand{\nZeroHitLargeShared}{3} %

\newcommand{\ci}[3]{#1 [#2, #3]}

\usepackage{authblk}

\title{Physical regime changes expose formula misuse in large language models}
\author[1]{Maohao Ran}
\author[2]{Yanting Zhang}
\author[1]{Chendong Ma}
\author[1,*]{Jun Song}
\affil[1]{Department of Geography, Hong Kong Baptist University, Hong Kong SAR, China}
\affil[2]{Hong Kong University of Science and Technology, Hong Kong SAR, China}
\affil[*]{Corresponding author: \texttt{junsong@hkbu.edu.hk}}
\date{\normalsize Code and data: \url{https://github.com/acodercat/AtmosRegime}\\[2pt]Trained models: \url{https://huggingface.co/collections/codercat/atmosregime}}

\begin{document}

\maketitle

\begin{abstract}
Large language models (LLMs) are increasingly employed for scientific computing across the Earth and environmental sciences, but problems in these fields often involve distinct physical regimes. LLMs may fail to recognize these regime-dependent constraints, applying formulas outside their domains of validity and producing plausible but physically incorrect results. We systematically investigate these failures through controlled experiments and assess whether post-training can improve regime-aware reasoning. Across 72 physical mechanisms, we find frontier LLMs frequently misapply formulas outside their regimes of validity, reducing accuracy by \insideMinusOutsideMin--\insideMinusOutsideMax{} percentage points. Reinforcement learning with verifiable rewards enables Qwen3.5-9B to match DeepSeek~V4~Flash (284B) on regime-dependent tasks, but its effectiveness is limited by the formulas available in the base model. Combining supervised fine-tuning on a few worked examples with reinforcement learning improves accuracy by \doseHOneRounded{} percentage points, while distillation from a teacher supplied with the correct formula raises accuracy from 0\% to \opdPRounded\%.
\end{abstract}

\section*{Introduction}

Large language models (LLMs) are increasingly integrated into scientific research, extending beyond information retrieval and question answering to experimental planning, programming, and quantitative analysis. LLM-based agents have demonstrated capabilities in designing and conducting experiments \citep{boiko2023coscientist,swanson2025virtuallab}, motivating broader efforts to incorporate these systems into scientific discovery \citep{wang2023discovery}. Earth and atmospheric sciences provide an important setting for these developments, as they rely extensively on quantitative relationships to interpret observations and describe physical processes across spatial and temporal scales. Machine learning has already become an important component of these fields, particularly in weather forecasting and the parameterization of unresolved processes in climate models \citep{reichstein2019,lam2023,bi2023,kochkov2024,eyring2024}. More recently, LLMs have been explored for domain-specific question answering, scientific programming, and quantitative calculations \citep{vaghefi2023,koldunov2024,song2025airgpt,zhou2026silent}, accompanied by benchmarks evaluating their capabilities in climate and atmospheric sciences \citep{manivannan2025climaqa,li2025atmossci}. As LLMs assume a more direct role in scientific computation, however, ensuring the physical validity of their outputs becomes increasingly important. A calculation may produce a numerically plausible, dimensionally consistent result, or even execute successfully as code, while relying on an inappropriate physical relationship. Such errors can be difficult to detect through routine inspection, particularly when apparently coherent model outputs are taken as evidence of correctness \citep{messeri2024,mittelstadt2023protect,zhou2024reliable}.

A fundamental challenge arises from the regime-dependent nature of physical laws and approximations. Many commonly used formulas are derived under specific assumptions about characteristic scales, dominant processes, or the magnitude of perturbations, and are valid only within the corresponding physical regimes. For example, the classical Stokes drag law assumes creeping flow in the continuum limit, whereas departures from these conditions require appropriate corrections or alternative drag formulations. Similarly, linear approximations may cease to be accurate when perturbations become sufficiently large for nonlinear effects to matter. Reliable scientific computation therefore requires not only evaluating formulas correctly, but also identifying their underlying assumptions, determining whether those assumptions hold, and selecting appropriate alternatives when they do not. The difficulty of generalizing across physical conditions is well recognized in machine learning, where parameterizations trained under limited environmental conditions can perform poorly in unseen regimes \citep{beucler2024}. Recent evaluations suggest analogous challenges for LLMs, which may continue to apply familiar formulas or solution strategies even when the physical regime requires a different approach \citep{ran2026atmoscoder,zhou2026silent,chung2025tpbench}. Importantly, such failures can arise from distinct limitations: a model may fail to recognize that the applicable regime has changed, or it may recognize the change but lack the ability to generate the required formula. These possibilities cannot be distinguished from an incorrect final answer alone, particularly given evidence that LLMs may exhibit apparent conceptual understanding without consistently applying the underlying principles \citep{mancoridis2025potemkin}. Distinguishing failures of regime recognition from limitations in formula availability is therefore essential for evaluating scientific reasoning in LLMs and understanding the extent to which post-training can address these limitations.

Whether post-training can overcome these distinct limitations also remains unclear. Domain-specific language models for geoscience, oceanography, and climate science have been developed through further training on specialized data \citep{deng2023k2,lin2024geogalactica,bi2024oceangpt,thulke2024climategpt}. However, acquiring domain knowledge does not necessarily translate into its reliable application: fine-tuning can be ineffective at incorporating new facts \citep{ovadia2024finetuning}, and models may reproduce learned knowledge without consistently applying it in reasoning tasks \citep{allenzhu2023knowledge,lampinen2025generalization}. Reinforcement learning with verifiable rewards (RLVR) offers an alternative by optimizing model behavior against objectively checkable outcomes, enabling smaller specialized models to approach the performance of substantially larger general-purpose models \citep{su2025rewardbridge}. Yet whether RLVR can extend a model's capabilities beyond those already exhibited by the base model remains debated. Some studies report improvements on previously unsolved problems \citep{deepseek2025,liu2025prorl}, whereas others suggest that RL primarily reinforces solutions already accessible to the base model \citep{yue2025,wu2025leash,matsutani2025squeezes}. Additional supervision through demonstrations, partial solutions, or teacher guidance can improve performance on problems that base models struggle to solve \citep{yan2025luffy,li2025questa,zhang2025scaf,nath2025guide,yan2026fest}. However, improvements in aggregate benchmark accuracy do not reveal whether training enhances the ability to select an appropriate solution or enables the model to generate previously inaccessible formulas. Regime-dependent scientific problems provide a setting in which these possibilities can be examined more directly, because the applicable formulas and their validity conditions are explicitly defined, their generation frequency can be measured before training, and the resulting calculations can be independently verified.

\begin{figure}[!ht]
\centering\includegraphics[width=\textwidth]{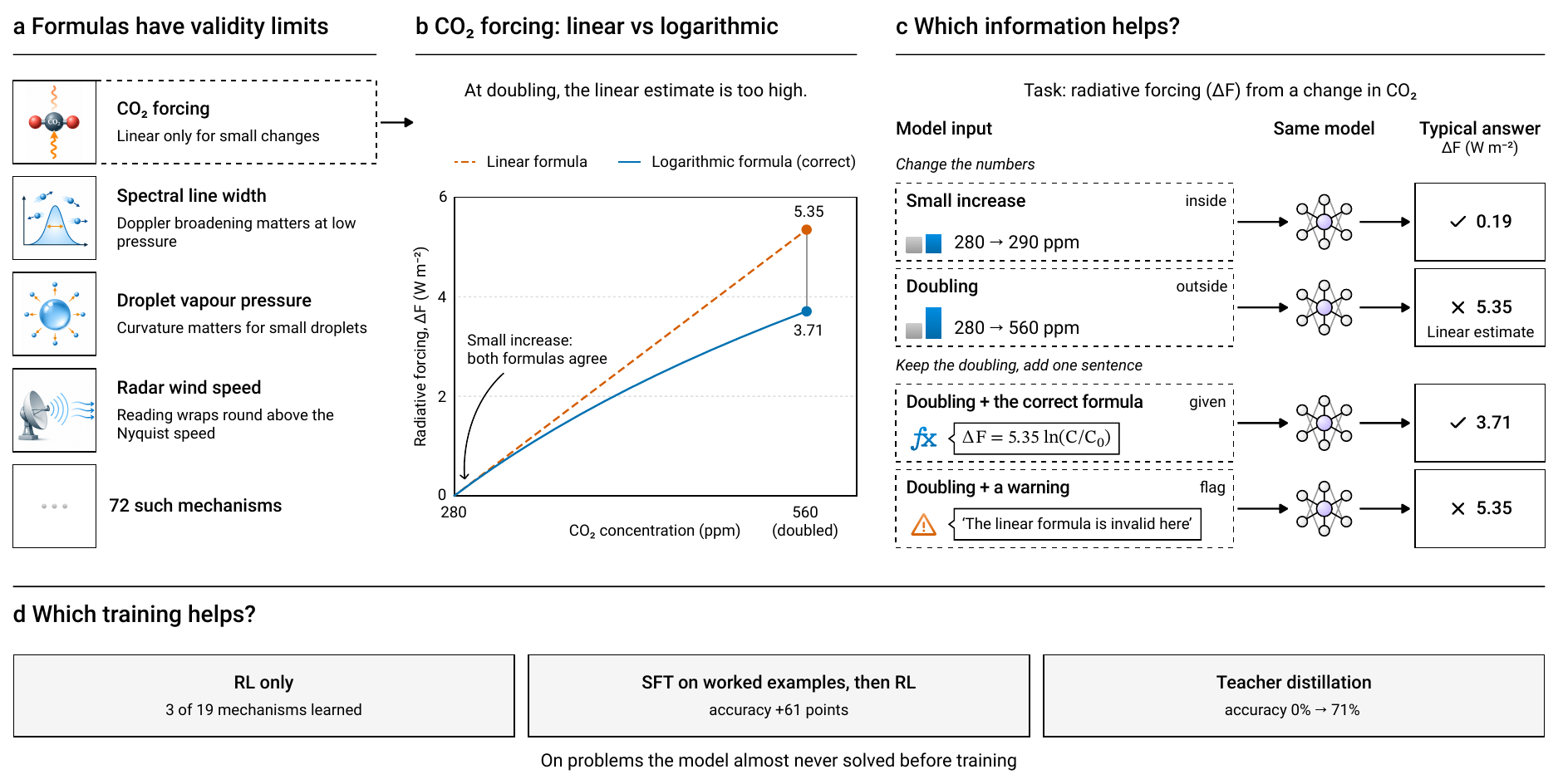}
\caption{\textbf{Physical regimes, controlled evaluation, and training strategies.}
\textbf{a}, Four representative mechanisms illustrating regime-dependent formula validity.
\textbf{b}, CO$_2$ radiative forcing as an example of regime-dependent approximations.
\textbf{c}, Four variants of the same problem testing formula applicability, regime warnings, and explicit formula provision. Model responses are illustrative, not measured frequencies.
\textbf{d}, Effects of three training strategies on mechanisms rarely solved by the base model: reinforcement learning (RL), supervised fine-tuning (SFT) followed by RL, and teacher-guided distillation (Figs.~\ref{fig:4}--\ref{fig:7}).}
\label{fig:concept}
\end{figure}

To investigate these questions, we construct a controlled benchmark covering \num{72} physical mechanisms from ten atmospheric subfields and related environmental sciences. Each mechanism is drawn from textbooks or primary literature, with its governing formulas and validity conditions verified against the original sources. We then construct four closely matched problem variants that differ by a single sentence (Fig.~\ref{fig:concept}): a problem within the standard formula's regime of validity, one outside that regime, one outside with the appropriate formula explicitly provided, and one outside with only a warning that the standard formula is inapplicable. For example, a radiative forcing problem contrasts a small increase in CO$_2$ concentration, where a linear approximation is appropriate, with a doubling that requires a different formulation. Comparing performance across these conditions allows us to assess whether models recognize changes in physical regimes and whether access to the correct formula resolves their errors. We evaluate \nLocalConfigs{} open-model configurations from five model families alongside frontier models, and quantify the discrepancy between standard-formula predictions and verified solutions across \nTrapsCost{} benchmark problems \citep{ran2026atmoscoder}.

Using this framework, we further investigate how different post-training strategies address these limitations. We first evaluate reinforcement learning on domain-specific problems with verifiable rewards \citep{yu2025dapo}. We then examine two approaches that introduce additional formula-related supervision: supervised fine-tuning on a few verified worked examples, assigned at random, before applying the same reinforcement learning procedure; and on-policy distillation using token-level guidance from a teacher model supplied with the correct formula. These experiments comprise \computeRuns{} reinforcement-learning runs across three models, totaling \computeGPUh{} GPU hours. All confirmatory analyses were pre-registered with explicit decision rules, and the outcomes of every rule, whether satisfied or not, are reported in Supplementary Table~1.

Our results reveal widespread failures to adapt familiar formulas across physical regimes. Providing the correct formula substantially reduces these errors, whereas simply warning that the standard formula is invalid offers limited improvement. Although reinforcement learning improves overall performance, it rarely resolves failures involving formulas that the base model seldom generates; targeted supervision through worked examples or teacher guidance can help overcome this limitation.

\section*{Results}

\subsection*{Models apply formulas outside their domains of validity}

\begin{figure}[!ht]
\centering\includegraphics[width=\textwidth]{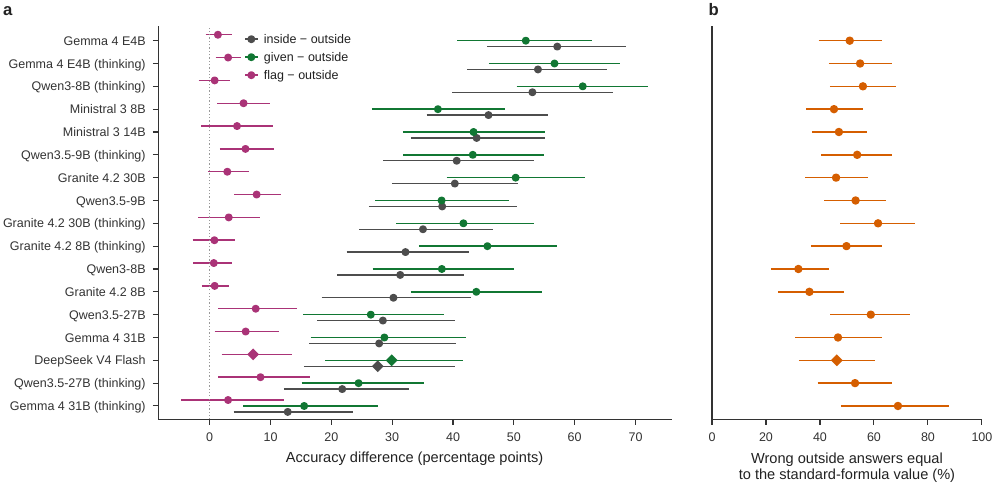}
\caption{\textbf{Regime-dependent accuracy losses and formula misapplication across models.}
Problem variants follow Fig.~\ref{fig:concept}; \emph{given}, correct formula provided; \emph{flag}, warning of formula invalidity only.
\textbf{a}, Accuracy differences across problem variants. Circles represent 16 open-model configurations; diamonds, DeepSeek~V4~Flash.
\textbf{b}, Proportion of incorrect out-of-regime answers within 2\% of the standard formula's prediction. Error bars indicate 95\% confidence intervals.}
\label{fig:1}
\end{figure}

Across \nLocalConfigs{} open-model configurations, accuracy on out-of-regime problems is \insideMinusOutsideMin{} to \insideMinusOutsideMax{} percentage points lower than on matched in-regime problems, with all corresponding 95\% intervals excluding zero (Fig.~\ref{fig:1}a; Methods). DeepSeek~V4~Flash exhibits a decline of \ci{\frontierInsideMinusOutside}{\frontierInsideMinusOutsideLo}{\frontierInsideMinusOutsideHi} percentage points, while GPT-5.5, GPT-5.4 and MiniMax-M3 show declines of \frontierTemplateGapMin{} to \frontierTemplateGapMax{} points with thinking disabled (Supplementary Note~1). Neither larger model size nor thinking consistently reduces this gap (Supplementary Fig.~1). A substantial fraction of incorrect out-of-regime answers closely matches the value predicted by the inapplicable standard formula: \shortcutShareMin--\shortcutShareMax\% of incorrect answers across open-model configurations, and \frontierShortcutShare\% for DeepSeek~V4~Flash, fall within 2\% of that value (Fig.~\ref{fig:1}b; Table~\ref{tab:mechanisms}). This pattern also appears in an independent set of 54 textbook problems. Modifications that invalidate the standard solution method produce an accuracy decline \ci{\dFiveGap}{\dFiveGapLo}{\dFiveGapHi} percentage points greater, relative to the original problems, than modifications that preserve its validity (Supplementary Fig.~2; Supplementary Table~2).

\begin{table}[!ht]
\caption{\textbf{Ten of the 72 mechanisms.} The inside and outside versions of each mechanism differ only in the triggering condition; the last column gives, for one outside problem, the value from the standard formula and the correct value. The 70 mechanisms of the training study, with base-model reachability and training outcomes, are in Supplementary Table~3.}
\label{tab:mechanisms}
\centering
\renewcommand{\arraystretch}{1.25}\setlength{\tabcolsep}{3pt}
\begin{tabular}{>{\raggedright\arraybackslash}p{2.3cm}>{\raggedright\arraybackslash}p{3.1cm}>{\raggedright\arraybackslash}p{3.1cm}>{\raggedright\arraybackslash}p{3.1cm}>{\raggedright\arraybackslash}p{3.0cm}}
\toprule
Subfield & Standard formula & Fails when & Replacement & Example: standard / correct \\
\midrule
Radiation & Line width $\propto p$ (Lorentz) & Low pressure; Doppler width dominates & Voigt width & 0.415 / 18.0 MHz \\
Radiation & Beer's law, $e^{-\tau}$ & Isotropic (diffuse) incident flux & Diffusivity factor, $e^{-1.66\tau}$ & 178 / 136 W\,m$^{-2}$ \\
Climate & Linear forcing, $S\,(C-C_0)$ & Concentration far from $C_0$ & $S C_0 \ln(C/C_0)$ & 6.72 / 4.35 W\,m$^{-2}$ \\
Thermodynamics & Linear extrapolation of $e_s$ & Large temperature change & Exponential (Clausius--Clapeyron) & 33.5 / 42.5 hPa \\
Aerosols & Stokes settling speed & Particle size near the mean free path & Cunningham slip correction & 4.16 / 7.16 \si{\micro\metre\per\second} \\
Chemistry & Henry's law, $H p$, fixed gas reservoir & Very soluble gas in a closed parcel & $H p/(1 + H R T L)$ & 41,600 / 19,400 nmol\,L$^{-1}$ \\
Cloud physics & Buoyancy from virtual temperature & Cloud carrying liquid water & Subtract condensate load $g\,r_l$ & 10.5 / 7.62 cm\,s$^{-2}$ \\
Observation \& modelling & Displayed = true radial velocity & Speed above the Nyquist velocity & Velocity folded into $[-v_N, v_N)$ & 21.0 / $-$7.88 m\,s$^{-1}$ \\
Hydrology & Melt $= Q/(\rho_w L_f)$ & Snowpack below \SI{0}{\celsius} & Cold content satisfied first & 38.9 / 28.2 mm \\
Cloud physics & Saturation vapour pressure over flat water & Drop radius of a few nanometres & Kelvin curvature factor & 9.35 / 11.7 hPa \\
\bottomrule
\end{tabular}
\end{table}

Stating the correct formula closes most of the gap. Outside accuracy is then within \givenMinusInsideMin{} to $+$\givenMinusInsideMax{} points of inside accuracy in every configuration, and the pre-registered $\pm$10-point bound holds for \sThreeMet{} of \sThreeN{} configurations on all 72 mechanisms (Supplementary Table~1). A warning that the standard formula does not hold, without the correct one, recovers \flagMinusOutsideMin{} to \flagMinusOutsideMax{} points.

\begin{figure}[!ht]
\centering\includegraphics[width=\textwidth]{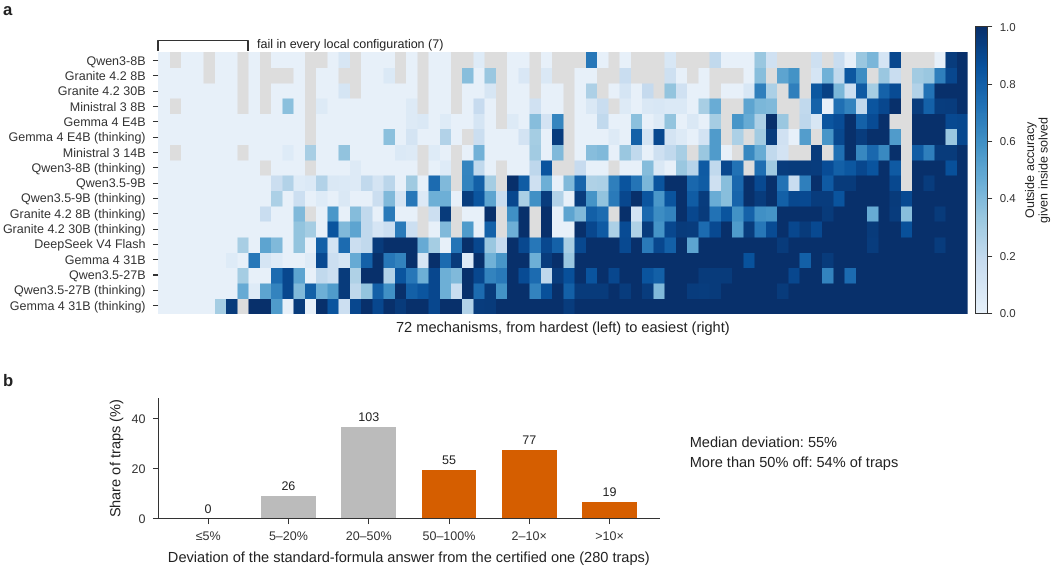}
\caption{\textbf{Distribution and magnitude of regime-dependent errors.}
\textbf{a}, Out-of-regime accuracy conditional on correct in-regime solutions, across 72 mechanisms and 17 model configurations ordered by overall performance. Grey indicates insufficient correct in-regime solutions.
\textbf{b}, Numerical deviation of standard-formula predictions from certified solutions across \nTrapsCostDefined{} benchmark traps. Numbers indicate problem counts.}
\label{fig:2}
\end{figure}

The same mechanisms fail in most models (Fig.~\ref{fig:2}a). Across 72 mechanisms and 17 configurations, the mechanism accounts for \atlasTemplateShare\% of the variance in outside accuracy (permutation $p = \atlasPermP$), and we detect no effect of subfield or of the type of approximation (Supplementary Note~2). \nUniversalFail{} mechanisms defeated every open-model configuration: each configuration that solved the inside version failed the outside version at least half the time (Supplementary Table~3). They are the square-root dependence of methane forcing on concentration, the exponential growth of saturation vapour pressure with temperature, the ventilation of an evaporating raindrop, the capacitance of a plate ice crystal, Doppler broadening of a spectral line at low pressure, the humidity correction of sonic-anemometer heat flux and the air mass of a sun photometer near the horizon. For a doubling of CO$_2$, extrapolating the forcing rate at 280 ppm linearly gives \coTwoDoublingLinear{} W\,m$^{-2}$ against the logarithmic \coTwoDoublingLog{}; at a solar zenith angle of 87$^\circ$, the plane-parallel air mass understates the optical depth retrieved from a sun photometer by \sunPhotometerDeviation\%.

To see how far off the standard method is, we took \nTrapsCost{} benchmark traps, textbook problems in which one condition of a parent problem was changed, and computed the answer the parent's method gives. The deviation from the certified answer is defined for \nTrapsCostDefined{} of them (the others change sign or are undefined). Its median is \costMedianDeviation\% [\costMedianDeviationLo, \costMedianDeviationHi]; the two answers differ by a factor of at least two for \costShareGeTwo\% of problems and by more than ten for \costCountGtTen{} (Fig.~\ref{fig:2}b). On the \costRegimeDefined{} regime-boundary traps, where the parent's method is a formula applied outside its regime, the median is \costRegimeMedian\% [\costRegimeMedianLo, \costRegimeMedianHi] and \costRegimeShare\% are more than 50\% off. A pressure-broadened line width computed at 0.1--0.4 hPa, where Doppler broadening dominates, is wrong by up to a factor of \lineBroadeningFactor{}. Subfields do not differ detectably in the share of large errors (Supplementary Fig.~3). These values are what a solver applying the parent's method would return, not the errors that models made, and the traps were built so that the two answers differ.

\subsection*{Supplying the formula beats prompting a regime check}

\begin{figure}[!ht]
\centering\includegraphics[width=\textwidth]{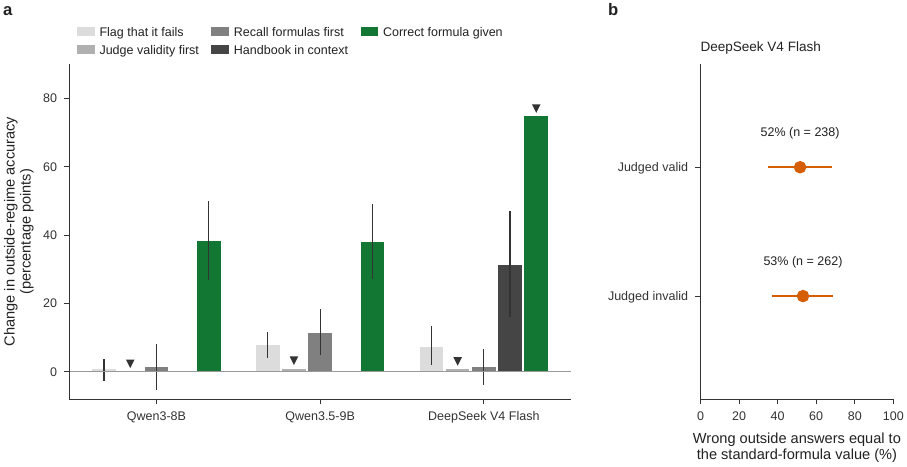}
\caption{\textbf{Five prompt interventions and the regime judgement.}
(a) Change in outside accuracy under five prompt interventions, each run on its own problem set (Supplementary Methods); triangles, no interval.
(b) Wrong outside answers of DeepSeek~V4~Flash that equal the standard-formula value, by whether the model judged the standard formula valid; $n$, wrong answers. Error bars, 95\% intervals.}
\label{fig:3}
\end{figure}

Prompting models to assess the validity of the standard formula has little effect on out-of-regime accuracy (Fig.~\ref{fig:3}a). When asked to determine whether the standard formula applies, models identify it as invalid in \judgeSaidNoQwen--\judgeSaidNoQwenB\% of out-of-regime problems. However, feeding these judgments back into their subsequent solutions changes overall accuracy by less than one percentage point. Even among problems explicitly identified as invalid, accuracy changes by \ci{\judgeTCQwenB}{\judgeTCQwenBLo}{\judgeTCQwenBHi} percentage points for Qwen3.5-9B and \ci{\judgeTCDS}{\judgeTCDSLo}{\judgeTCDSHi} percentage points for DeepSeek~V4~Flash. Prompting models to recall relevant formulas before solving similarly changes out-of-regime accuracy by \ci{\recallQwen}{\recallQwenLo}{\recallQwenHi} percentage points for Qwen3-8B and \ci{\recallQwenB}{\recallQwenBLo}{\recallQwenBHi} percentage points for Qwen3.5-9B. In contrast, providing explicit formula information yields larger improvements. On 17 mechanisms, supplying a handbook containing 60 formulas, including the applicable one, increases DeepSeek~V4~Flash accuracy from \handbookNone\% to \handbookWith\%. Providing only the applicable formula raises accuracy on the same problems to \handbookGiven\%.

The models often say that the formula fails and then use it anyway. Asked alongside each problem whether the standard formula holds, DeepSeek~V4~Flash answers ``no'' for \detectH\% of outside problems and \detectF\% of inside problems (difference \ci{\detectHminusF}{\detectHminusFLo}{\detectHminusFHi} points), so its judgement carries information. Yet among its wrong outside answers, the share equal to the standard formula's value is the same whether it had judged the formula valid (\judgedYesShortcut\%) or invalid (\judgedNoShortcut\%) (Fig.~\ref{fig:3}b).

\subsection*{Reinforcement learning rarely repairs unsolved mechanisms}

\begin{figure}[!ht]
\centering\includegraphics[width=\textwidth]{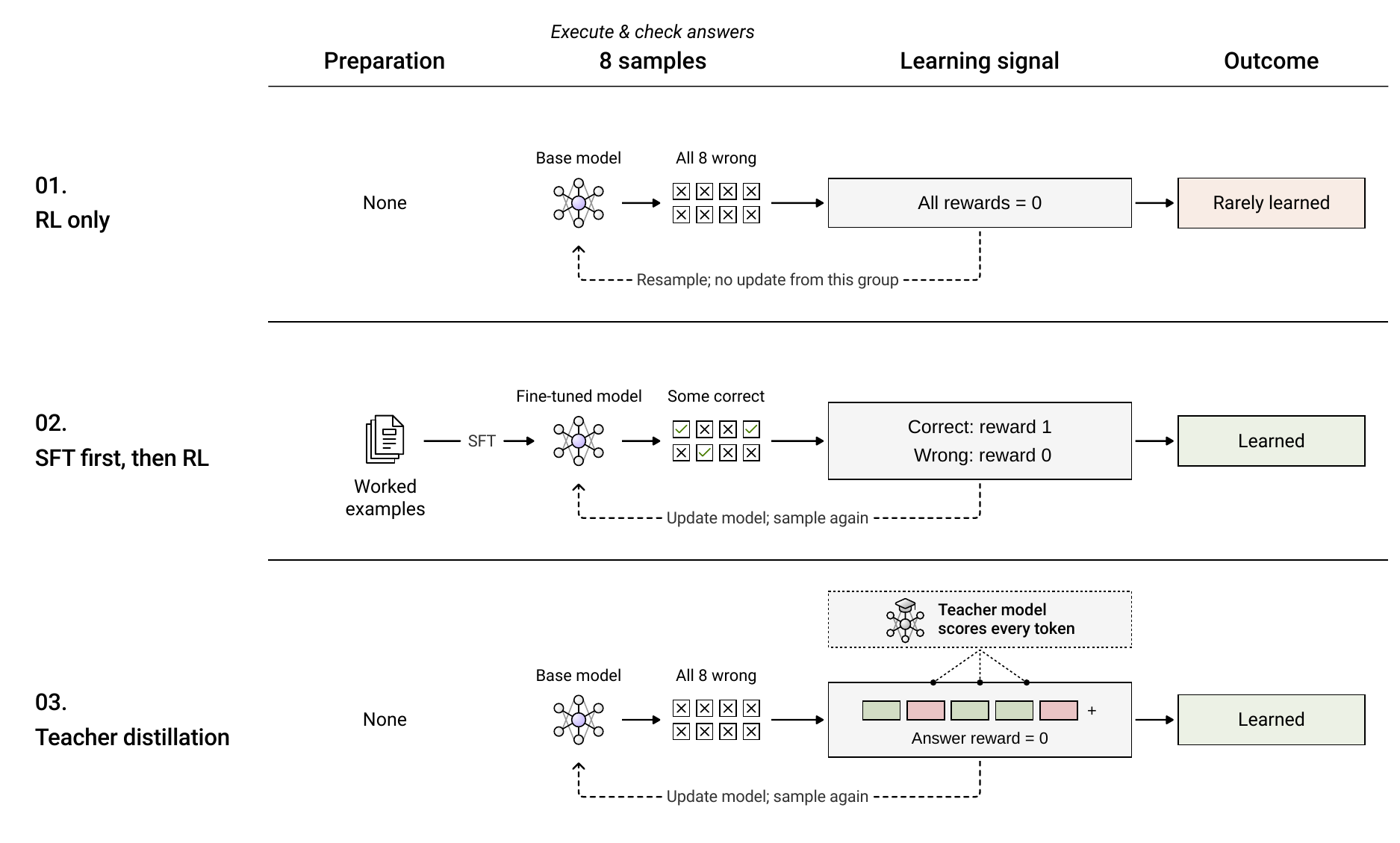}
\caption{\textbf{Three training routes for a mechanism the base model cannot solve.} Each row is one route; the model always sees the problem without the formula. The patterns of right and wrong programs are illustrative, not measured rates. RL, reinforcement learning; SFT, supervised fine-tuning.}
\label{fig:4}
\end{figure}

\begin{figure}[!ht]
\centering\includegraphics[width=\textwidth]{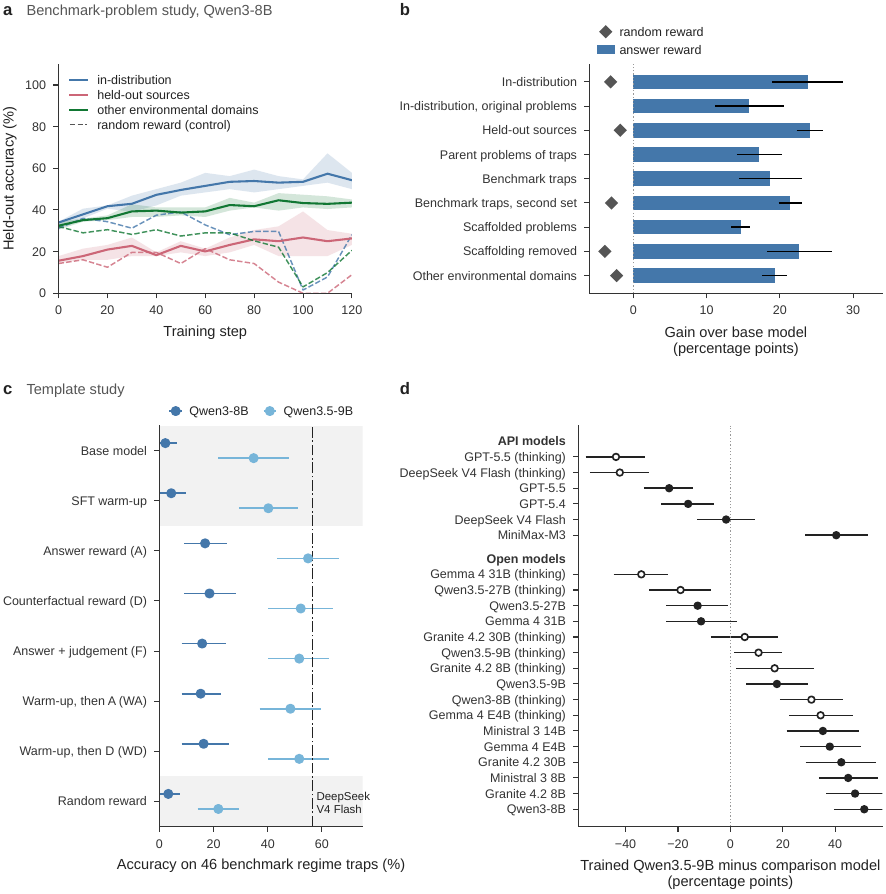}
\caption{\textbf{Reinforcement learning on domain problems generalises and lifts a 9B model to the level of DeepSeek~V4~Flash on regime traps.}
(a, b) Qwen3-8B trained on 190 benchmark problems: held-out accuracy during training (a; mean and range of four runs) and gain over the base model on nine held-out probe sets (b; mean $\pm$ s.d. of four runs; diamonds, random reward).
(c) Accuracy on \nRegimeTraps{} benchmark regime traps after training on the templates; dash-dotted line, DeepSeek~V4~Flash without thinking; shaded rows, models before reinforcement learning and the random-reward control.
(d) Trained Qwen3.5-9B minus each comparison model, paired over the traps both were run on; filled, thinking off; open, thinking on. Error bars in c and d, 95\% intervals.}
\label{fig:5}
\end{figure}

We ran two training studies without the formula in the prompt. The first trains on benchmark problems and measures how far the gains carry. The second trains on the mechanisms themselves and compares reinforcement learning alone with two ways of adding the formula (Fig.~\ref{fig:4}). In the first study, Qwen3-8B was trained by reinforcement learning with verifiable rewards (DAPO \citep{yu2025dapo}; eight sampled programs per problem, reward 1 for a correct executed answer) on \num{190} atmospheric benchmark problems and their numeric variants. Accuracy rose on problems the model had not trained on (Fig.~\ref{fig:5}a,b): by \rlGainTestId{} points on held-out problems of the same distribution, \rlGainTestBook{} on problems from two held-out sources, \rlGainScaffoldStripped{} on problems with their scaffolding removed and \rlGainCrossDomain{} on \nCrossDomain{} problems from hydrology, soil science, environmental chemistry and ecology, in line with reports that outcome-reward reinforcement learning generalises to unseen variants of a task \citep{chu2025sftrl}. All \nProbeRuns{} measurements (nine problem sets, four runs) were positive. With a random reward the same training moved the nine sets by between \randomRewardNineMin{} and $+$\randomRewardNineMax{} points, so the gains do not come from the spurious-reward effect reported for some model families \citep{shao2025spurious}.

These averages do not show which mechanisms were repaired, so the second study follows mechanisms one at a time. The 70 mechanisms were split by cluster into two folds, and each model was trained on one fold and evaluated on the other, in both directions. Five training variants share the problems and differ in reward or starting point: an answer reward; a counterfactual reward, which also runs the program on the matching inside problem; an answer reward plus a reward for a correct regime judgement; and the first two after a warm-up, one epoch of supervised fine-tuning (SFT) on the base model's own verified solutions. A control uses a random reward (arms A, D, F, WA, WD and R in Methods). Qwen3-8B (two seeds) and Qwen3.5-9B were trained (training dynamics in Supplementary Fig.~4). We score regime judgement by $H - F$, the share of outside problems the model declares invalid minus the share of inside problems it declares so.

Training on the mechanisms raises accuracy on benchmark regime traps, which require the solver to notice that the standard method fails. On \nRegimeTraps{} held-out traps, Qwen3-8B rises from \trapsBaseQwen\% to \ci{\trapsAQwen}{\trapsAQwenLo}{\trapsAQwenHi}\% and Qwen3.5-9B from \trapsBaseQwenB\% to \ci{\trapsAQwenB}{\trapsAQwenBLo}{\trapsAQwenBHi}\% (Fig.~\ref{fig:5}c). On the same traps and with thinking off, the trained Qwen3.5-9B matches DeepSeek~V4~Flash, \frontierAQwenB\% against \frontierDSOff\% (paired difference \ci{\frontierDiffDS}{\frontierDiffDSLo}{\frontierDiffDSHi} points), and exceeds MiniMax-M3 (\frontierMiniMaxOff\%). GPT-5.5 and GPT-5.4 with thinking off (\frontierGPTFiveFiveOff\% and \frontierGPTFiveFourOff\%) and every frontier model with thinking on (\frontierOnMin--\frontierOnMax\%) remain above it. It also exceeds every open-model configuration of 14B or less and closes \gapClosedLarge\% of the gap to Qwen3.5-27B (Fig.~\ref{fig:5}d; Supplementary Table~4). None of the compared models was trained on the mechanisms. The gain holds on an independent benchmark set. After training, Qwen3-8B improves by \ci{\indepRegimeGain}{\indepRegimeGainLo}{\indepRegimeGainHi} points on its \indepRegimeN{} regime traps and by at least as much on the geometry traps and parent problems, and with a random reward it does not improve (Supplementary Table~4).

On the held-out fold, Qwen3-8B's outside accuracy rises by only \heldGainQwenMin{} to \heldGainQwenMax{} points across training variants, less than its inside accuracy (outside minus inside gain under the answer reward \ci{\asymHeldQwenA}{\asymHeldQwenALo}{\asymHeldQwenAHi} points). The pre-registered transfer criterion, a held-out rise of at least 15 points, is met by Qwen3.5-9B in \transferArmsMetQwenB{} of five variants and not by Qwen3-8B (Supplementary Table~1). The judgement reward does teach the regime judgement, and the effect carries to mechanisms never trained on. Held-out $H - F$ is \ci{\pTwoFminusAQwen}{\pTwoFminusAQwenLo}{\pTwoFminusAQwenHi} points higher than under the answer reward for Qwen3-8B and \ci{\pTwoFminusAQwenBSeed}{\pTwoFminusAQwenBSeedLo}{\pTwoFminusAQwenBSeedHi} for Qwen3.5-9B.

\begin{figure}[!ht]
\centering\includegraphics[width=\textwidth]{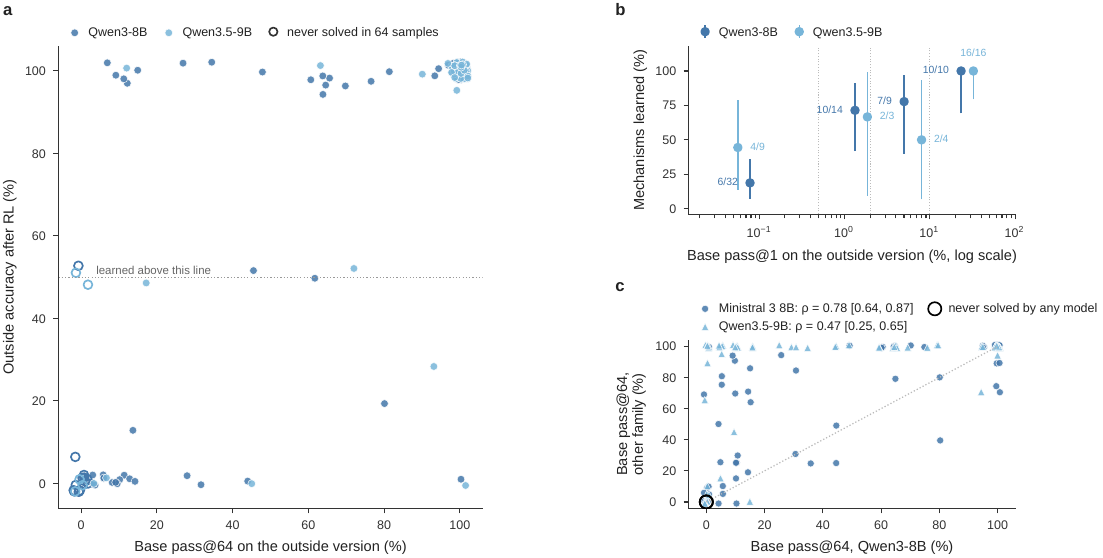}
\caption{\textbf{Reinforcement learning rarely repairs what the base model never solves.}
(a) Base pass@64 against accuracy after training, per mechanism; open, never solved in 64 samples; points are jittered.
(b) Share of mechanisms learned against how often the base model solved them (pass@1, binned; labels, learned/total).
(c) Base pass@64 per mechanism in three models; circled, never solved by any. Error bars in b, 95\% intervals.}
\label{fig:6}
\end{figure}

The base model's success rate predicts which mechanisms reinforcement learning learns. Before training we drew 64 samples per outside problem from the base model: pass@64 is the share of problems solved at least once and pass@1 the share of samples that are correct, and we refer to both as reachability. Qwen3-8B produced no correct answer in \samplesPerMechanism{} samples for \nZeroHit{} mechanisms, and \nZeroHitNotLearnedAny{} of these stayed unlearned in every training variant without warm-up and in every seed (Fig.~\ref{fig:6}a). Qwen3.5-9B had \nZeroHitQwenB{} such mechanisms (seed 42) and learned \nZeroHitLearnedQwenB{} of them in every variant. Across both models, \nZeroHitLearnedPooled{} of \nZeroHitPooled{} never-solved mechanisms were learned. Base pass@64 predicts which mechanisms training repairs with an AUC of \ci{\pOneAUC}{\pOneAUCLo}{\pOneAUCHi} (Fisher $p = \pOneFisherP$, never-solved against the rest, 69 mechanisms; Supplementary Table~1).

Some of the rare correct samples were right by coincidence. Rerun with new outside parameters, only about \luckFunctionalRare\% of them still gave the correct answer, and counting only these strengthens the association (Fisher $p = \luckFisherP$). For Qwen3-8B, the share of mechanisms learned rises from \pcurveQwenBinZeroLearned/\pcurveQwenBinZeroN{} where fewer than 0.5\% of base samples are correct to \pcurveQwenBinThreeLearned/\pcurveQwenBinThreeN{} above 10\% (Fig.~\ref{fig:6}b; all four bins in Supplementary Note~3). The steepest part of this rise lies where a group of eight samples begins to contain both right and wrong answers, which is the condition for a non-zero policy gradient and has probability $1-(1-p)^8-p^8$ at pass@1 $= p$. The Qwen3.5-9B bins, with three and four mechanisms in the middle, are not monotone.

Which mechanisms a base model can solve is partly shared across models. Base pass@64 per mechanism correlates between Qwen3-8B and a third base model, Ministral~3~8B (Spearman $\rho$ = \ci{\rhoMinQwen}{\rhoMinQwenLo}{\rhoMinQwenHi}) and between Qwen3-8B and Qwen3.5-9B (\ci{\rhoQwenQwenB}{\rhoQwenQwenBLo}{\rhoQwenQwenBHi}; 69 mechanisms each), and \zeroInAllFamilies{} mechanisms were never solved by any of the three base models (Fig.~\ref{fig:6}c). Qwen3.5-27B, three times the size of Qwen3.5-9B, ranks the mechanisms much as the 9B model does ($\rho$ = \ci{\rhoQwenBLarge}{\rhoQwenBLargeLo}{\rhoQwenBLargeHi}) and never solves \nZeroHitLarge{} of them, \nZeroHitLargeShared{} of which the 9B model also misses.

Neither a different reward nor more training moved the limit set by the base model. The counterfactual reward adds \ci{\counterfactualGain}{\counterfactualGainLo}{\counterfactualGainHi} points over the answer reward on the training fold. We continued one answer-reward run of Qwen3-8B for \extSteps{} further steps on \extMechs{} mechanisms it had not learned, sampling each problem about ten times as often as before. None was learned. The number with zero pass@64 rose from \extZeroPassStart{} to \extZeroPassEnd{}, and inside accuracy stayed at 100\%. For \extNoSignal{} of the \extMechs{}, none of roughly \num{8500} outside samples per mechanism was rewarded, so the policy gradient on them stayed at zero. A warm-up on the base model's own solutions, written with the formula stated and then paired with the plain problem, does add \ci{\warmupGain}{\warmupGainLo}{\warmupGainHi} points, an effect the next experiment tests with a randomised design.

\subsection*{Worked examples let reinforcement learning repair more}

\begin{figure}[!ht]
\centering\includegraphics[width=\textwidth]{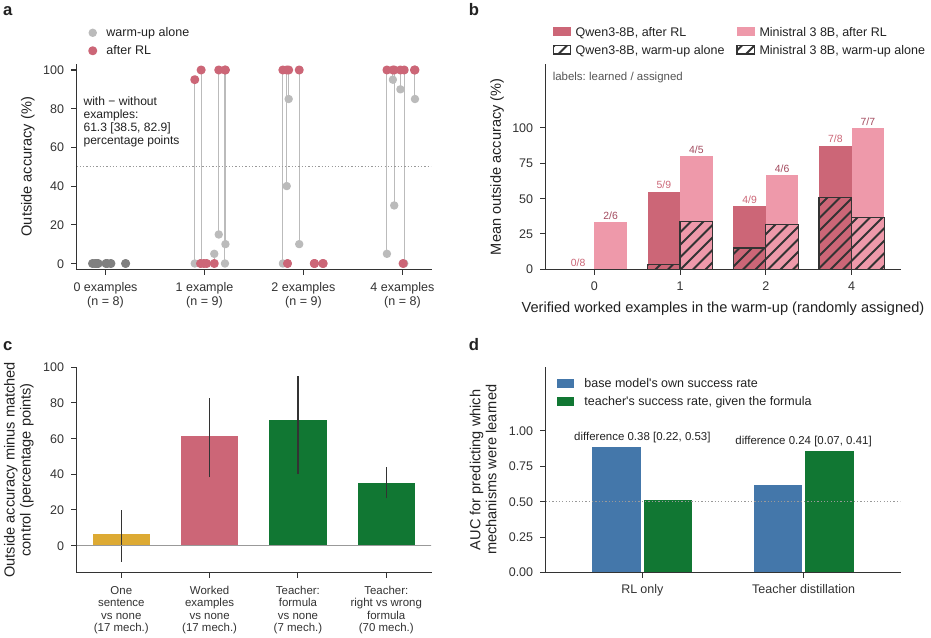}
\caption{\textbf{Verified worked examples repair mechanisms that reinforcement learning alone could not.}
(a) Outside accuracy of Qwen3-8B after warm-up (grey) and after reinforcement learning (red; dark grey without examples), by the randomly assigned number of worked examples.
(b) Mean outside accuracy by number of examples, two models; hatched, warm-up alone.
(c) Effect of four ways of supplying the formula, each against its matched control.
(d) How well the student's and the teacher's starting accuracy predict which mechanisms each method learned. Error bars in c, 95\% intervals.}
\label{fig:7}
\end{figure}

To determine whether targeted supervision can overcome failures that persist after reinforcement learning (RL), rather than merely reflect differences in mechanism difficulty, we randomised the number of worked examples provided before RL. We selected \doseMechanisms{} mechanisms that were rarely solved by the base model (pass@64 $\leq 0.1$) and remained below 50\% accuracy after RL alone in both training seeds. Each example consisted of a solver program generated by the base model when supplied with the correct formula, verified by execution, and paired with the original out-of-regime problem and its in-regime counterpart. The correct formula was therefore introduced into training only through the program. For each mechanism and seed, we randomly assigned 0, 1, 2 or 4 examples during supervised warm-up, while holding all other training conditions and the subsequent RL procedure constant (Fig.~\ref{fig:7}a,b).

Providing worked examples increased final out-of-regime accuracy by \ci{\doseHOne}{\doseHOneLo}{\doseHOneHi} percentage points relative to no examples (randomisation test, $p = \doseRandP$; Supplementary Fig.~5). Of the \doseQwenDosedN{} mechanism–seed observations assigned examples, \doseQwenDosedLearned{} were learned, compared with \doseQwenZeroLearned{} of \doseQwenZeroN{} without examples. Among six mechanisms assigned examples in only one of the two seeds, four achieved approximately 100\% accuracy with examples but 0\% without, while none showed the reverse pattern. Importantly, most of the improvement emerged during subsequent RL rather than supervised warm-up alone. With one example, warm-up accuracy was \doseQwenDOneWarm\%, increasing to \doseQwenDOneRL\% after RL; with four examples, the corresponding accuracies were \doseQwenDFourWarm\% and \doseQwenDFourRL\%. Across all observations receiving examples, RL increased out-of-regime accuracy by an additional \ci{\doseRLminusWarm}{\doseRLminusWarmLo}{\doseRLminusWarmHi} percentage points, while only \doseWarmOnlyLearned{} of \doseQwenDosedN{} observations reached 50\% accuracy after warm-up alone.

A replication on Ministral~3~8B (12 mechanisms, three of them controls without examples) went in the same direction but fell short of the pre-registered criterion. The difference between observations with and without examples is \ci{\doseMinHOne}{\doseMinHOneLo}{\doseMinHOneHi} points; one control mechanism was learned from the generic warm-up alone (Supplementary Note~4). All \doseMinDFourN{} observations given four examples were learned, as were \doseMinDosedLearned\% of all observations with examples. When both models are pooled, with a threshold chosen after seeing the data, every observation whose warm-up model solved the outside version in at least 2\% of samples was learned (\pooledHighPLearned{} of \pooledHighPN).

Supplying the same formula as a one-sentence statement instead of a program changed trained accuracy by \ci{\doseTextHOne}{\doseTextHOneLo}{\doseTextHOneHi} points in a within-mechanism crossover. The warm-up models never reproduced the sentence when asked, and its content rarely appeared in their solutions (Supplementary Note~5).

The formula can also come from a teacher, and the base model can be its own, because for \opdMainSet{} mechanisms it never solved the outside problems but did solve them when the formula was stated. We trained Qwen3-8B on its own samples with token-by-token guidance from a frozen copy of itself whose prompt contained the correct formula (privileged on-policy distillation; arms P, S and Pshuf in Methods). On these mechanisms, outside accuracy rose from 0 to \ci{\opdP}{\opdPLo}{\opdPHi}\% (two seeds, the same five mechanisms learned). With a teacher given no formula, outside accuracy reached \opdS\% (difference \ci{\opdPminusS}{\opdPminusSLo}{\opdPminusSHi}), and with a teacher shown a different mechanism's formula, \opdPshuf\% (difference \ci{\opdPminusPshufMain}{\opdPminusPshufMainLo}{\opdPminusPshufMainHi}); across all 70 mechanisms, the wrong formula left accuracy \ci{\opdPminusPshufAll}{\opdPminusPshufAllLo}{\opdPminusPshufAllHi} points below the right one (Fig.~\ref{fig:7}c). The distillation runs used no dynamic sampling and are therefore compared with one another, not with the answer-reward runs (Methods).

Over the \sTwoUnits{} mechanisms the base model does not solve (Fig.~\ref{fig:7}d), the student's own pass@64 predicts which ones reinforcement learning repairs (AUC \aucStudentA) and the teacher's accuracy when given the formula does not (\aucTeacherA; difference \ci{\aucDiffA}{\aucDiffALo}{\aucDiffAHi}). For distillation the pattern reverses (\aucStudentP{} against \aucTeacherP; difference \ci{\aucDiffP}{\aucDiffPLo}{\aucDiffPHi}).

\subsection*{Accuracy improves but regime discrimination does not}

\begin{figure}[!ht]
\centering\includegraphics[width=\textwidth]{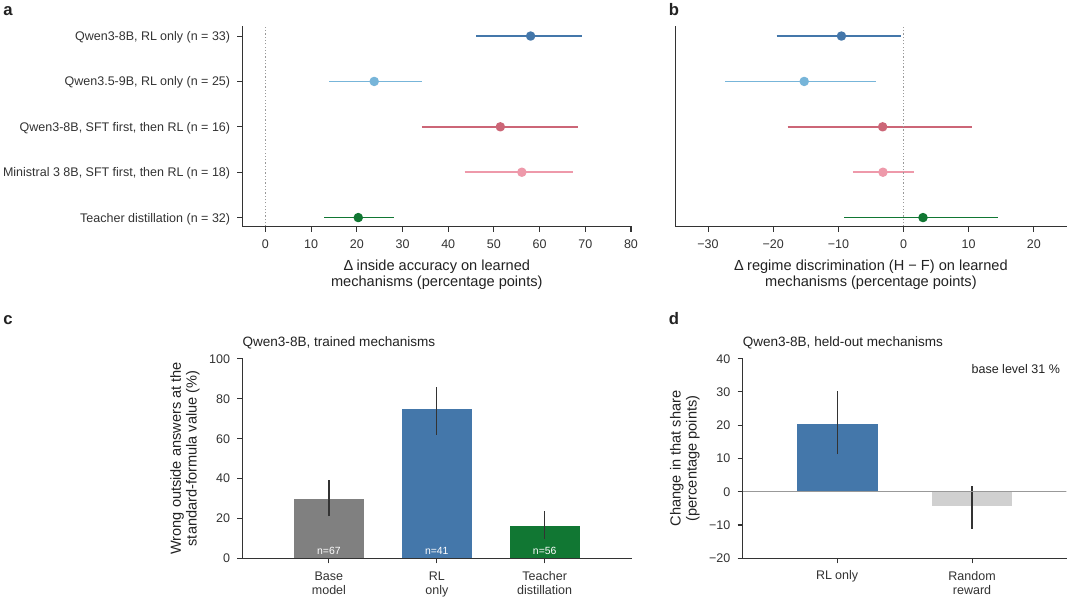}
\caption{\textbf{What is learned and what stays wrong.}
(a) Change in inside accuracy on the mechanisms each condition learned.
(b) Change in regime discrimination ($H - F$) on the same mechanisms.
(c) Wrong outside answers that equal the standard-formula value after each route (Qwen3-8B; $n$, mechanisms).
(d) Change in that share on the held-out fold. Error bars, 95\% intervals.}
\label{fig:8}
\end{figure}

Across nine training conditions spanning three models, we identified \rFourLearnedCells{} mechanism--condition pairs with successful out-of-regime learning. In none of these pairs did in-regime accuracy decline beyond the prespecified 25-percentage-point margin (\rFourSwapCells{} cases; largest decline, \rFourMaxInsideDrop{} points; Fig.~\ref{fig:8}a shows five conditions). For mechanisms learned by Qwen3-8B under answer-reward training, in-regime accuracy instead increased by \ci{\rFourDinsideQwenA}{\rFourDinsideQwenALo}{\rFourDinsideQwenAHi} percentage points. Notably, the models had not consistently solved these in-regime problems before training. Among mechanisms subsequently learned, baseline in-regime and out-of-regime accuracies were \rFourInsideStartQwenA\% and \rFourOutsideStartQwenA\% for Qwen3-8B under answer-reward training, \rFourInsideStartQwenDose\% and \rFourOutsideStartQwenDose\% for the worked-example condition, \rFourInsideStartP\% and \rFourOutsideStartP\% for distillation, and \rFourInsideStartQwenBA\% and \rFourOutsideStartQwenBA\% for Qwen3.5-9B. After training, in-regime accuracy reached \rFourInsideEndP--\rFourInsideEndQwenDose\%, indicating improvements in solving problems even when the standard formula remained applicable.

However, improved computational accuracy did not translate into more reliable regime discrimination. On the same mechanisms, answer-reward training changed the $H-F$ contrast by \ci{\rFourDhfQwenA}{\rFourDhfQwenALo}{\rFourDhfQwenAHi} percentage points for Qwen3-8B and \ci{\rFourDhfQwenBA}{\rFourDhfQwenBALo}{\rFourDhfQwenBAHi} percentage points for Qwen3.5-9B (Fig.~\ref{fig:8}b). These changes indicate reduced discrimination between the two regimes despite improvements in numerical accuracy. For models trained with worked examples or distillation, the corresponding confidence intervals included zero.

Evidence that formulas introduced through worked examples transfer to other problems remains inconclusive. On \nRegimeTraps{} benchmark regime traps, models trained with examples achieved \rFourDoseTraps\% accuracy, compared with \rFourWATraps\% under otherwise matched training without examples (difference \ci{\rFourDoseMinusWA}{\rFourDoseMinusWALo}{\rFourDoseMinusWAHi}; Supplementary Fig.~6). Distilled models remained below 10\% accuracy. Because these benchmark problems were not matched to the mechanisms receiving worked examples, the aggregate comparison cannot establish whether the introduced formulas transfer to corresponding unseen problems. One traced case is presented in Supplementary Note~6. The generalization gains reported in Fig.~\ref{fig:5}c arise from reinforcement learning itself.

Finally, incorrect predictions after training continued to cluster around values given by the inapplicable standard formulas. On the training fold, the proportion of incorrect out-of-regime answers matching the standard formula's prediction was \shortcutLevelbase\% for the base model, \shortcutLevelA\% after answer-reward training, and \shortcutLevelP\% after distillation (Fig.~\ref{fig:8}c). These proportions involve different sets of mechanisms with remaining errors across training conditions. On the held-out fold, the corresponding proportion was \shortcutBaseLevel\% for the base model and changed by \ci{\shortcutHeldA}{\shortcutHeldALo}{\shortcutHeldAHi} percentage points after answer-reward training and by \ci{\shortcutHeldR}{\shortcutHeldRLo}{\shortcutHeldRHi} percentage points under random-reward training (Fig.~\ref{fig:8}d). Nevertheless, incorrect predictions from trained models were numerically closer to the verified solutions than those from the base model (Supplementary Fig.~7).

\section*{Discussion}

Reliable scientific computation requires not only evaluating formulas correctly, but also determining whether they apply under the relevant physical conditions. Our results reveal a persistent limitation in this distinction: across a wide range of physical mechanisms, LLMs frequently apply familiar formulas outside their regimes of validity, producing plausible but incorrect results. These failures are not consistently eliminated by increasing model size or enabling additional reasoning, and they differ from errors involving the inconsistent execution of otherwise appropriate formulas \citep{ran2026atmoscoder}. Notably, prompting models to assess a formula's validity provides little improvement, even when they explicitly identify the formula as inapplicable. In contrast, supplying the correct alternative formula recovers much of the lost accuracy. This contrast suggests that recognizing a regime violation is insufficient for successful correction, while access to and effective use of the applicable formula constitutes an important constraint. It does not, however, establish whether the missing formula is absent from the model's internal knowledge or merely inaccessible during problem solving.

Our training experiments provide further evidence that the effectiveness of reinforcement learning depends on the solution capabilities exhibited by the base model. Reinforcement learning with verifiable rewards substantially improves overall accuracy, enabling a 9B model to match a much larger frontier model on regime-dependent problems under comparable non-thinking settings. Yet these gains are concentrated in mechanisms that the base model can already solve at least occasionally. Mechanisms with little or no initial success remain difficult to learn, even under prolonged, targeted training or alternative reward formulations. The probability of successful learning rises sharply when initial samples contain a mixture of correct and incorrect solutions, consistent with reinforcement learning strengthening solution strategies already accessible to the model. Similar relationships between initial success and subsequent learning have been reported in medical calculations and synthetic training settings \citep{lin2025ehrmind,zhang2025interplay}. These findings inform the broader debate over whether reinforcement learning extends model capabilities or primarily improves the reliability of existing ones \citep{yue2025,liu2025prorl,wu2025leash,matsutani2025squeezes}. In our setting, the evidence favors the latter interpretation for mechanisms rarely solved by the base model, although failure to observe a correct solution in a finite sample does not establish that the model lacks the underlying knowledge.

Importantly, the mechanisms resistant to reinforcement learning alone are not necessarily intrinsically unlearnable. By randomly assigning small numbers of verified worked examples before otherwise identical reinforcement learning, we find that targeted supervision can substantially improve performance on previously resistant mechanisms. Even a single example enabled learning in \doseOneLearned{} of \doseOneN{} observations, and much of the improvement emerged during the subsequent reinforcement learning rather than supervised warm-up alone. The examples were executable solver programs generated by the base model when provided with the correct formula, allowing the relevant information to enter training without a separate teacher model. This approach resembles earlier efforts to bootstrap reasoning from generated solutions \citep{zelikman2022star} and complements methods that use hints, solution prefixes, demonstrations, or other forms of guidance to improve learning on initially unsolved problems \citep{yan2025luffy,li2025questa,zhang2025scaf,nath2025guide,ma2026relift,yan2026fest}. On-policy distillation from a teacher supplied with the correct formula also improved previously difficult mechanisms, whereas guidance from a teacher given an incorrect formula was substantially less effective. These observations suggest that the content of additional supervision can matter when the required solution is rarely generated by the student, although studies on mathematical reasoning have reported weaker effects of reference correctness and stronger effects of demonstration structure \citep{shrestha2026privileged,li2025structure}. The comparison of guidance content remains tentative because of its limited mechanism coverage and single-seed controls.

These findings have practical implications for developing domain-specific language models and scientific agents. Rather than relying exclusively on aggregate benchmark accuracy, training strategies can be informed by mechanism-level evaluations of what a base model can already solve. Initial success rates estimated from repeated sampling provide an empirical indication of which failures may respond to reinforcement learning alone, while teacher performance when supplied with the correct formula can help assess the potential of distillation. For mechanisms with consistently low initial success, verified worked examples or formula-guided supervision may offer a more effective starting point. Importantly, providing a formula as a declarative statement during training showed no clear benefit: models rarely reproduced the statement or used its content in their solutions. This experiment therefore cannot distinguish a failure to acquire the information from a failure to apply it, consistent with previous findings that learned factual knowledge may not be reliably deployed in reasoning \citep{allenzhu2023knowledge,ruis2024procedural,lampinen2025generalization}. For domain models developed through additional training on geoscience, oceanographic, or climate-related corpora \citep{deng2023k2,lin2024geogalactica,bi2024oceangpt,thulke2024climategpt}, evaluation should therefore test whether formulas can be applied under the appropriate physical conditions, rather than whether they can merely be recalled. This is particularly relevant for relationships such as CO$_2$ and methane radiative forcing and the Clausius--Clapeyron relation, which underpin important climate calculations but were among the mechanisms that the trained base models could not reliably solve.

Improved computational performance should nevertheless not be equated with improved awareness of physical validity. On mechanisms successfully learned during training, out-of-regime accuracy increased without a corresponding deterioration in in-regime accuracy beyond the prespecified margin; in some cases, in-regime performance also improved. However, answer-reward training reduced the models' ability to distinguish the applicability of formulas across regimes as measured by explicit validity judgments. Furthermore, remaining incorrect answers frequently clustered around the predictions of the inapplicable standard formula, even as their numerical values moved closer to the correct solutions. Because the mechanisms contributing incorrect answers differ across training conditions, these changes in error distributions do not establish that individual failures became more severe or more difficult to detect. Nevertheless, they illustrate a limitation of optimizing scientific reasoning solely through final-answer correctness: improved numerical accuracy need not imply more reliable assessment of the assumptions underlying a calculation. This distinction is related to broader concerns about outcome-based training signals and the reliability of model judgments \citep{kalai2025,song2025tax,kirichenko2025abstention,damani2025rlcr}. For scientific applications, evaluating both sides of a regime boundary is therefore essential, particularly when apparently reasonable numerical outputs may conceal the use of an invalid physical approximation.

Several limitations constrain the generality of these conclusions. Training experiments were conducted on 8--9B parameter models, comprising \computeRuns{} runs, whereas larger models were evaluated but not trained. Comparisons with frontier models used non-thinking settings; with thinking enabled, frontier models reached \frontierOnMin--\frontierOnMax\% accuracy on the same regime traps, substantially above the trained models. The randomized worked-example experiment included 17 mechanisms evaluated across two seeds, with a replication involving 12 mechanisms and three controls; the resulting intervals remain wide. Our benchmark also emphasizes relatively uncommon formulas and uses controlled templates. Although evaluation instances contain parameter values absent from training, examples and evaluation problems share template structures, so transfer to independently constructed problems involving the same physical mechanisms remains untested. The independent textbook evaluations demonstrate that regime-dependent failures extend beyond the original templates, but do not establish corresponding generalization of the training interventions. Moreover, because training includes problems from both sides of a regime boundary, successful performance does not establish whether a model has learned to select formulas conditionally or has adopted a more general formula across both regimes. Analyses developed after the primary results are exploratory and are identified as such in Methods.

As LLM-based agents assume greater responsibility for scientific computation, ensuring physical validity will require more than verifying the numerical execution of their solutions. Our results suggest a mechanism-level approach in which formula applicability is evaluated explicitly, persistent failures are identified through controlled tests, and targeted supervision supplies verified solution procedures where reinforcement learning alone is insufficient. Such interventions are traceable to particular physical mechanisms and examples, offering a practical basis for auditing and improving domain-specific models. A critical next step is to establish whether these improvements transfer beyond the templates used in training to independently sourced problems involving the same mechanisms, and whether models can reliably distinguish when different physical formulations are required. Ultimately, dependable scientific reasoning requires not only producing accurate answers, but also applying the physical relationships that make those answers valid.

\section*{Methods}

\subsection*{Pre-registration}
Each experiment was written down with its decision rules and committed before the runs it governs; later changes were appended as dated amendments. The pre-registration documents and their dated records are supplied with the code, and the outcome of every confirmatory rule is listed in Supplementary Table~1.

\subsection*{Problems}
\textbf{Mechanisms and templates.} The 72 mechanisms were selected by the authors from standard textbooks and the primary literature, and for each the standard formula, its validity limit and the replacement formula were checked by hand against the source by one author and independently by a second. They fall in 32 clusters (a cluster joins mechanisms that share a standard formula, a correction or a solving habit) and cover ten atmospheric subfields and neighbouring environmental sciences. Each template draws parameters and emits four versions of one problem: \emph{inside}, where the standard formula holds; \emph{outside}, where one triggering quantity is moved across the regime boundary so that it fails; \emph{given}, the outside text plus one sentence stating the correct formula; and \emph{flag}, the outside text plus one sentence stating that the standard treatment does not hold, without saying what replaces it. Templates were accepted only if the shortcut and certified answers agreed within 2\% inside and differed by at least 15\% outside, every quantity the replacement needs appears in all versions, and the replacement is never named in the outside or flag text; each solver was re-derived independently and checked against a literature anchor. The diagnostic set draws 20 parameter pairs per template (seed 20261008; 5,760 rows). The training study uses 70 templates (two excluded beforehand on physics grounds), split by cluster into two folds (F1: 29 templates in 16 clusters; F2: 41 in 16). Training pools hold 20 instances per template (seed 20260930). The final evaluation set, 70 templates $\times$ 20 pairs $\times$ 4 versions (seed 20261010; 5,600 rows), was drawn after the pools, used for no parameter update or checkpoint choice, and evaluated on the base model (starting accuracy and pass@64) before training. Three templates (line broadening, drop ventilation, Clausius--Clapeyron extrapolation) were reworded for the training study after the diagnostic run, so the local-configuration contrasts are reported over the other \nTemplatesDiag{} templates (\nClustersDiag{} clusters). Including the three widens inside $-$ outside in every configuration (by at most \rewordMaxChange{} points; Supplementary Table~2) and changes no pre-registered verdict, so the reported values are conservative. Other diagnostic analyses use the original templates.

\textbf{Benchmark problems and traps.} Parent problems are the 436 core problems of AtmosCoder-Bench \citep{ran2026atmoscoder}, drawn from textbooks and frozen into a training split (190 problems, five numeric variants each, 1,140 rows) and test splits (64 in-distribution, 63 core-only, 56 from two held-out textbooks). Traps change one condition of one parent so that the parent's method no longer applies and store the parent's (shortcut) answer; the regime-recognition family requires the solver to notice the failure and is compared with geometry-detail traps that change a detail without invalidating the method. The diagnostic rule used the 38 test-side regime traps available at the time against the geometry traps; the training-time regime set has 46 test-side traps, of which the 38 are a subset; an independent replication used 54 regime traps sharing none with the first. The cost analysis uses the 328 traps with a stored shortcut answer and no magnitude defect (Supplementary Table~5 lists every problem set and its size); the 131 cross-domain problems come from hydrology, soil science, environmental chemistry and ecology. In the cost analysis, deviations up to 50\% are binned by $|s - c|/|c|$ and larger ones by $\max(s/c, c/s)$, where $s$ is the standard-formula answer and $c$ the certified one.

\subsection*{Diagnostic evaluation}
Sixteen configurations from five families (Qwen3-8B; Qwen3.5 9B and 27B; Gemma~4 E4B and 31B; Granite~4.2 8B and 30B; each with thinking off and on; Ministral~3 8B and 14B, off only) were served with vLLM 0.24 (bfloat16, 32,768-token context). DeepSeek~V4~Flash (a mixture-of-experts model with 284B parameters, 13B active per token) was evaluated through its API with thinking disabled. Sampling used temperature 0.3, top-p 1.0 and at most 8,192 tokens (28,672 with thinking); templates were run once per configuration and benchmark traps three times.

The model writes a \texttt{solve()} program, which is executed in a sandbox and graded per sub-answer at 5\% relative tolerance with unit-aware comparison; an ungradable output (no code, or code that fails) is fed back for repair, up to five attempts. The pre-registered rule for a configuration was inside $-$ outside $\ge 20$ points on the templates and a paired regime $-$ geometry gap $\ge 15$ points on the benchmark traps, both intervals excluding zero. The shortcut share counts wrong outside answers within 2\% of the standard-formula value.

The detection probe adds an instruction to report, as a key, whether the standard treatment is valid; answers are re-executed with the key removed; $H$ is the share of outside rows judged invalid, $F$ the share of inside rows judged so, and a configuration ``can detect'' if $H - F \ge 20$ points with the interval above zero on at least 80\% parsable rows. Judge-first, recall-first and handbook interventions are described in Supplementary Methods.

\subsection*{Reinforcement learning}
Qwen3-8B (seeds 42, 43) and Qwen3.5-9B (seed 42; seed 43 added after the main results and reported with it), thinking off, were trained with DAPO in verl (commit 5b79827b; vLLM 0.24, torch 2.11, transformers 5.9) on eight H800 GPUs per run: learning rate $10^{-6}$, 128 prompts $\times$ 8 rollouts per step in four optimiser updates of 32 prompts, clip 0.2/0.28, no KL or entropy term, token-mean loss, dynamic sampling on the score (groups with identical rewards are discarded and re-sampled, at most ten extra batches), maximum response 3,072 tokens, no overlong penalty, checkpoints every 5 steps. The reward is 1 if the executed program's answer matches the certified answer under a hardened unit rule; a failed execution counts only if it fails again on an immediate rerun. For a problem $x$ with sampled programs $y_1,\dots,y_8$ and rewards $r_i \in \{0,1\}$, the loss is
\begin{equation}
\begin{aligned}
\hat A_i&=\frac{r_i-\bar r}{s_r+10^{-6}},\qquad
\rho_{i,t}=\frac{\pi_\theta(y_{i,t}\mid x,y_{i,<t})}{\pi_{\theta_{\mathrm{old}}}(y_{i,t}\mid x,y_{i,<t})},\\
\mathcal L_{\mathrm{RL}}(\theta)&=-\frac{1}{\sum_i |y_i|}\sum_i\sum_{t=1}^{|y_i|}\min\!\Big(\rho_{i,t}\hat A_i,\ \operatorname{clip}(\rho_{i,t},\,0.8,\,1.28)\,\hat A_i\Big),
\end{aligned}
\label{eq:rl}
\end{equation}
where $\bar r$ and $s_r$ are the mean and sample standard deviation of the eight rewards, the sums run over every response token of a mini-batch (32 problems $\times$ 8 programs), groups whose eight rewards are identical are discarded and re-sampled, and negative advantages carry the dual-clip bound of 3 that verl applies by default.

Per fold, five arms share problems and order: the answer reward (arm A); a counterfactual reward (arm D: 0 if the answer is wrong, 0.5 if right, 1 if the same program, re-executed with the triggering parameter set to the inside twin's value, also solves the twin); the answer reward plus 0.25 for a correct regime judgement (arm F, implemented as weights 0.8 : 0.2; only F's template rows carry the judgement instruction); and the answer and counterfactual rewards started from a warm-up (arms WA and WD). The warm-up samples each given row of the training fold eight times (temperature 1), keeps the first program correct on both inside and outside versions, pairs it with the plain inside and outside text, mixes it with the base model's own verified solutions to standard problems (at most two per row). It then runs one epoch of supervised fine-tuning (learning rate $5\times10^{-6}$, batch 32, maximum length 4,096, loss on the assistant turn only, seed 42; loss curves in Supplementary Fig.~8).

The random-reward control (arm R) replaces the reward by an independent fair coin per rollout (correctness logged, never rewarded) and was added after the main results. A fixed 256-row training-fold probe is evaluated every 20 steps for the training curves only.

A run ends at 120 steps, at 24 hours of wall clock, when its step has not advanced for 2 hours (sampling saturation), or after a second crash (one same-seed resume allowed). The five-arm Qwen3-8B runs took 4--12 hours and reached 120 steps (random-reward runs \rArmHoursQwenMin--\rArmHoursQwenMax{} hours). Qwen3.5-9B runs took 12--24 hours (random reward \rArmHoursQwenBMin--\rArmHoursQwenBMax{} hours), and some were stopped by the cap or the saturation rule between steps 88 and 116 (Supplementary Fig.~4).

The first study (Fig.~\ref{fig:5}a,b; training reward in Supplementary Fig.~9) used the same recipe on the 190 benchmark training problems (main run plus three seeds, 120 steps; random-reward control at the same step), evaluated with the benchmark harness on nine held-out probes, among them the problems from two textbooks held out whole.

Base pass@$k$ uses 64 samples per outside row of the final set (temperature 1.0, top-p 1.0, 8,192 tokens, no repair), averaged per template; sensitivity at pass@32, temperature 0.7, is in Supplementary Note~3. Endpoints are evaluated on the 5,600 final rows with the diagnostic protocol (one sample per row); accuracy uses a re-execution of each response with any judgement key removed; rows with no response after retries enter no denominator; a response is truncated if it used all 8,192 tokens (0.5--1.0\% of rows for Qwen3-8B arms, 0.1--3.9\% for Qwen3.5-9B). Training-fold accuracy is an arm's accuracy on its own fold, held-out accuracy on the other.

Three rules were pre-registered. The reachability rule holds if no never-solved mechanism (base pass@64 $= 0$) reaches 50\% in any arm and seed and Spearman(pass@64, outside accuracy) $\ge 0.5$ with interval above zero in every arm (one mechanism excluded beforehand because its inside pass@64 was already 30\%). The transfer rule holds if held-out outside accuracy rises $\ge 15$ points with interval above zero. The regime-judgement rule holds if the F $-$ A difference in held-out $H - F$ is $\ge 20$ points with interval above zero, both arms $\ge 80\%$ parsable and F within 3 points of A on accuracy; for a family whose base model already detects the regime (Qwen3.5-9B), the pre-registration replaced the first condition by F $-$ start $\ge 15$ points with interval above zero, F $-$ A being reported alongside. A mechanism is counted as learned when its trained outside accuracy reaches 50\%; analyses that also require a start below 50\% say so.

\subsection*{Randomised examples}
In the randomised-example study, units are the training-fold mechanisms of Qwen3-8B with base pass@64 $\le 0.10$, A-arm outside accuracy $< 0.5$ in both seeds and at least four verified pairs available (17 mechanisms). Separately for each fold and reinforcement-learning seed, levels 0, 1, 2 and 4 were assigned by a seeded generator, $\lfloor n/4 \rfloor$ units per level with the remainder drawn without replacement (34 observations: 8, 9, 9 and 8). An example is one executable solver sampled from the base model on a given row (the outside problem with the correct formula stated), verified correct on both versions, and paired with the hint-free outside text and its inside twin (two rows); the mechanism's own warm-up pairs are drawn without replacement by a seeded generator. The warm-up data are those of the WA arm with each unit's pairs cut to its dose (byte-identical to WA at full dose); the training that follows is identical to WA.

The pre-registered criterion was a difference in mean trained outside accuracy between units with and without examples of at least 30 points, with a lower bound above zero (mechanism bootstrap, 10,000 draws). A mechanism is learned if its trained outside accuracy is $\ge 50\%$. The randomisation test re-draws the assignment within fold$\times$seed 10,000 times and counts draws with a difference at least as large as observed.

The text arm replaces each unit's examples by four question--answer rows whose answer is the template's given sentence, copied verbatim, in a within-mechanism crossover (each mechanism injected in exactly one seed); its criterion is the paired difference under the same rule. Ministral~3~8B (thinking off) repeated the design on 12 mechanisms (24 observations) with a warm-up built from its own verified solutions; its four runs reached the 24-hour cap at steps 90--105.

\subsection*{Privileged on-policy distillation}
In privileged on-policy distillation, the student is Qwen3-8B; the teacher is the frozen base model shown, for each outside row, the given twin of the problem. The privileged arm (P) adds to the loss of equation~\ref{eq:rl} a forward KL over the teacher's 32 most probable tokens at every position of the student's own programs,
\begin{equation}
\begin{aligned}
d_{i,t}&=\sum_{v\in\mathcal V_{i,t}}\pi_T(v\mid x^{+},y_{i,<t})\,\log\frac{\pi_T(v\mid x^{+},y_{i,<t})}{\pi_\theta(v\mid x,y_{i,<t})},\\
\mathcal L_{\mathrm P}(\theta)&=\mathcal L_{\mathrm{RL}}(\theta)+\frac{1}{\sum_i |y_i|}\sum_{i,t}\min\!\big(10,\ \max(0,\,d_{i,t})\big),
\end{aligned}
\label{eq:opd}
\end{equation}
where $\pi_T$ is the frozen base model, $x^{+}$ the teacher's prompt and $\mathcal V_{i,t}$ the teacher's top-32 tokens. The distillation term is back-propagated directly through $\pi_\theta$ rather than through the policy gradient \citep{agarwal2024gkd}. In P, $x^{+}$ is the given twin; the same-prompt control (S) shows the teacher the student's prompt ($x^{+} = x$, no privilege); the wrong-formula control (Pshuf) shows it a different mechanism's formula (a seeded derangement). Pk3 replaces $d_{i,t}$ by the $k_3$ estimator on the sampled token, $\operatorname{clip}(e^{\Delta}-\Delta-1,\,-10,\,10)$ with $\Delta=\log\pi_T(y_{i,t}\mid x^{+},y_{i,<t})-\log\pi_\theta(y_{i,t}\mid x,y_{i,<t})$, and Pk2 by $\tfrac12\Delta^2$ with the per-token clamp at 10, which all other arms share, switched off.

Training otherwise follows arm A (120 steps, 128 $\times$ 8 rollouts, mini-batches of 32) but with dynamic sampling off (its effect is examined in Supplementary Note~7), the student on four GPUs and the teacher on one. Because arm A ran with dynamic sampling on, the pre-registered P $-$ A contrast is confounded by the pipeline; the P $-$ S and P $-$ Pshuf contrasts, within the distillation pipeline, carry the result. The main test set is the seven mechanisms with base pass@64 $= 0$ whose given-row accuracy in the warm-up samples was $\ge 0.5$. The pre-registered rules required P $-$ A, P $-$ S and P $-$ Pshuf each to be at least 30 points with a lower bound above zero. All arms used seed 42, and P was repeated with seed 43.

An arm with a larger external teacher (Qwen3-32B or Qwen3-30B-A3B) was pre-registered but gated on a probe: shown the plain outside problems, the teacher had to solve at least \teacherGate{} of the \teacherProbeZ{} never-solved mechanisms at pass@1 $\ge 0.5$. Qwen3-32B solved \teacherKnownLarge{} and Qwen3-30B-A3B \teacherKnownMoe{}, so the arm was not run.

\subsection*{Statistics}
Units of independence are mechanisms, clusters or traps, never rows or repetitions; templates are weighted equally within a cluster and clusters equally; seeds are averaged within a unit before resampling; in the randomised experiments the two seed observations of a mechanism are kept separate, and bootstrap samples draw mechanisms, keeping both observations of each drawn mechanism.

Frontier and open models were compared with the trained model in the training study's evaluation harness (same prompt, temperature 0.3, 8,192 output tokens or 28,672 with thinking, 5\% tolerance, three repetitions). The frontier models were GPT-5.5 and GPT-5.4 (reasoning effort none), MiniMax-M3 and DeepSeek~V4~Flash (thinking disabled), and, as references with thinking, DeepSeek~V4~Flash (reasoning effort high) and GPT-5.5 (reasoning effort medium). The frontier models were run through their APIs on all \frontierN{} regime traps and their parent problems; the open configurations come from the diagnostic study's local runs, which cover \sharedN{} of these traps. Qwen3.5-27B was run on the remaining traps as well. The share of the gap closed is the trained model's gain over its base model divided by the 27B model's lead over that base model. The independent benchmark set (\indepRegimeN{} regime traps, \indepGeometryN{} geometry traps and \indepParentN{} parent problems; no trap in common with the set above) was run once per problem on Qwen3-8B trained on one fold with one seed. Qwen3.5-9B was not run on it. Paired differences between two models resample the traps both were run on (10,000 draws).

Intervals are percentile intervals from 10,000 resamples of units with a fixed, named generator per analysis. Learned-share intervals in Fig.~\ref{fig:6}b are Clopper--Pearson. Spearman and AUC intervals resample mechanisms. Fisher's exact tests are two-sided; permutation and randomisation tests are one-sided and count the draws at least as large as the observed value.

The analyses of what is learned, the three-model comparison, the transfer to benchmark problems, the fraction learned by base success rate, the pooled reachability threshold, the restriction of the diagnostic contrasts to 69 templates and the comparisons with frontier and open models on the benchmark traps were specified after the main results and are exploratory. These post-hoc analyses had their readings written before computation and were run twice to byte-identical output.

\section*{Data availability}
The templates, benchmark problems, train/test partition and result files used in this study are available at \url{https://github.com/acodercat/AtmosRegime}. One trained model for each of the three training routes is available at \url{https://huggingface.co/collections/codercat/atmosregime}. Model outputs and the other checkpoints are available from the corresponding author on reasonable request.

\section*{Code availability}
The code, pre-registration documents and scripts that reproduce the figures are available at \url{https://github.com/acodercat/AtmosRegime}.

\bibliographystyle{naturemag}
\bibliography{refs}

\begin{thebibliography}{10}
\expandafter\ifx\csname url\endcsname\relax
  \def\url#1{\texttt{#1}}\fi
\expandafter\ifx\csname urlprefix\endcsname\relax\def\urlprefix{URL }\fi
\providecommand{\bibinfo}[2]{#2}
\providecommand{\eprint}[2][]{\url{#2}}

\bibitem{boiko2023coscientist}
\bibinfo{author}{Boiko, D.~A.}, \bibinfo{author}{MacKnight, R.},
  \bibinfo{author}{Kline, B.} \& \bibinfo{author}{Gomes, G.}
\newblock \bibinfo{title}{Autonomous chemical research with large language
  models}.
\newblock \emph{\bibinfo{journal}{Nature}} \textbf{\bibinfo{volume}{624}},
  \bibinfo{pages}{570--578} (\bibinfo{year}{2023}).

\bibitem{swanson2025virtuallab}
\bibinfo{author}{Swanson, K.}, \bibinfo{author}{Wu, W.},
  \bibinfo{author}{Bulaong, N.~L.}, \bibinfo{author}{Pak, J.~E.} \&
  \bibinfo{author}{Zou, J.}
\newblock \bibinfo{title}{The {Virtual} {Lab} of {AI} agents designs new
  {SARS-CoV-2} nanobodies}.
\newblock \emph{\bibinfo{journal}{Nature}} \textbf{\bibinfo{volume}{646}},
  \bibinfo{pages}{716--723} (\bibinfo{year}{2025}).

\bibitem{wang2023discovery}
\bibinfo{author}{Wang, H.} \emph{et~al.}
\newblock \bibinfo{title}{Scientific discovery in the age of artificial
  intelligence}.
\newblock \emph{\bibinfo{journal}{Nature}} \textbf{\bibinfo{volume}{620}},
  \bibinfo{pages}{47--60} (\bibinfo{year}{2023}).

\bibitem{reichstein2019}
\bibinfo{author}{Reichstein, M.} \emph{et~al.}
\newblock \bibinfo{title}{Deep learning and process understanding for
  data-driven {Earth} system science}.
\newblock \emph{\bibinfo{journal}{Nature}} \textbf{\bibinfo{volume}{566}},
  \bibinfo{pages}{195--204} (\bibinfo{year}{2019}).

\bibitem{lam2023}
\bibinfo{author}{Lam, R.} \emph{et~al.}
\newblock \bibinfo{title}{Learning skillful medium-range global weather
  forecasting}.
\newblock \emph{\bibinfo{journal}{Science}} \textbf{\bibinfo{volume}{382}},
  \bibinfo{pages}{1416--1421} (\bibinfo{year}{2023}).

\bibitem{bi2023}
\bibinfo{author}{Bi, K.} \emph{et~al.}
\newblock \bibinfo{title}{Accurate medium-range global weather forecasting with
  {3D} neural networks}.
\newblock \emph{\bibinfo{journal}{Nature}} \textbf{\bibinfo{volume}{619}},
  \bibinfo{pages}{533--538} (\bibinfo{year}{2023}).

\bibitem{kochkov2024}
\bibinfo{author}{Kochkov, D.} \emph{et~al.}
\newblock \bibinfo{title}{Neural general circulation models for weather and
  climate}.
\newblock \emph{\bibinfo{journal}{Nature}} \textbf{\bibinfo{volume}{632}},
  \bibinfo{pages}{1060--1066} (\bibinfo{year}{2024}).

\bibitem{eyring2024}
\bibinfo{author}{Eyring, V.} \emph{et~al.}
\newblock \bibinfo{title}{Pushing the frontiers in climate modelling and
  analysis with machine learning}.
\newblock \emph{\bibinfo{journal}{Nature Climate Change}}
  \textbf{\bibinfo{volume}{14}}, \bibinfo{pages}{916--928}
  (\bibinfo{year}{2024}).

\bibitem{vaghefi2023}
\bibinfo{author}{Vaghefi, S.~A.} \emph{et~al.}
\newblock \bibinfo{title}{{ChatClimate}: grounding conversational {AI} in
  climate science}.
\newblock \emph{\bibinfo{journal}{Communications Earth \& Environment}}
  \textbf{\bibinfo{volume}{4}}, \bibinfo{pages}{480} (\bibinfo{year}{2023}).

\bibitem{koldunov2024}
\bibinfo{author}{Koldunov, N.} \& \bibinfo{author}{Jung, T.}
\newblock \bibinfo{title}{Local climate services for all, courtesy of large
  language models}.
\newblock \emph{\bibinfo{journal}{Communications Earth \& Environment}}
  \textbf{\bibinfo{volume}{5}}, \bibinfo{pages}{13} (\bibinfo{year}{2024}).

\bibitem{song2025airgpt}
\bibinfo{author}{Song, J.}, \bibinfo{author}{Ma, C.} \& \bibinfo{author}{Ran,
  M.}
\newblock \bibinfo{title}{{AirGPT}: pioneering the convergence of
  conversational {AI} with atmospheric science}.
\newblock \emph{\bibinfo{journal}{npj Climate and Atmospheric Science}}
  \textbf{\bibinfo{volume}{8}}, \bibinfo{pages}{179} (\bibinfo{year}{2025}).

\bibitem{zhou2026silent}
\bibinfo{author}{Zhou, T.}, \bibinfo{author}{Qian, Y.} \&
  \bibinfo{author}{Leung, L.~R.}
\newblock \bibinfo{title}{Can we trust {LLMs} for complex {Earth} system model
  analysis? {Silent} failure and evidence from module-grounded benchmarking}.
\newblock \emph{\bibinfo{journal}{EGUsphere [preprint]}}
  (\bibinfo{year}{2026}).
\newblock \bibinfo{note}{Preprint posted 27 April 2026; under review for
  Geoscientific Model Development}.

\bibitem{manivannan2025climaqa}
\bibinfo{author}{Manivannan, V.~V.} \emph{et~al.}
\newblock \bibinfo{title}{{ClimaQA}: an automated evaluation framework for
  climate question answering models}.
\newblock In \emph{\bibinfo{booktitle}{International Conference on Learning
  Representations}} (\bibinfo{year}{2025}).
\newblock \bibinfo{note}{{arXiv}:2410.16701}.

\bibitem{li2025atmossci}
\bibinfo{author}{Li, C.}, \bibinfo{author}{Deng, W.}, \bibinfo{author}{Lu, M.}
  \& \bibinfo{author}{Yuan, B.}
\newblock \bibinfo{title}{{AtmosSci-Bench}: evaluating the recent advance of
  large language model for atmospheric science}.
\newblock \bibinfo{howpublished}{Preprint at
  \url{https://arxiv.org/abs/2502.01159}} (\bibinfo{year}{2025}).

\bibitem{messeri2024}
\bibinfo{author}{Messeri, L.} \& \bibinfo{author}{Crockett, M.~J.}
\newblock \bibinfo{title}{Artificial intelligence and illusions of
  understanding in scientific research}.
\newblock \emph{\bibinfo{journal}{Nature}} \textbf{\bibinfo{volume}{627}},
  \bibinfo{pages}{49--58} (\bibinfo{year}{2024}).

\bibitem{mittelstadt2023protect}
\bibinfo{author}{Mittelstadt, B.}, \bibinfo{author}{Wachter, S.} \&
  \bibinfo{author}{Russell, C.}
\newblock \bibinfo{title}{To protect science, we must use {LLMs} as zero-shot
  translators}.
\newblock \emph{\bibinfo{journal}{Nature Human Behaviour}}
  \textbf{\bibinfo{volume}{7}}, \bibinfo{pages}{1830--1832}
  (\bibinfo{year}{2023}).

\bibitem{zhou2024reliable}
\bibinfo{author}{Zhou, L.} \emph{et~al.}
\newblock \bibinfo{title}{Larger and more instructable language models become
  less reliable}.
\newblock \emph{\bibinfo{journal}{Nature}} \textbf{\bibinfo{volume}{634}},
  \bibinfo{pages}{61--68} (\bibinfo{year}{2024}).

\bibitem{beucler2024}
\bibinfo{author}{Beucler, T.} \emph{et~al.}
\newblock \bibinfo{title}{Climate-invariant machine learning}.
\newblock \emph{\bibinfo{journal}{Science Advances}}
  \textbf{\bibinfo{volume}{10}}, \bibinfo{pages}{eadj7250}
  (\bibinfo{year}{2024}).

\bibitem{ran2026atmoscoder}
\bibinfo{author}{Ran, M.} \emph{et~al.}
\newblock \bibinfo{title}{Execution-grounded evaluation reveals hidden failures
  in language-model calculations for environmental science}.
\newblock \bibinfo{howpublished}{Preprint at
  \url{https://arxiv.org/abs/2608.18726}} (\bibinfo{year}{2026}).

\bibitem{chung2025tpbench}
\bibinfo{author}{Chung, D. J.~H.} \emph{et~al.}
\newblock \bibinfo{title}{Theoretical {Physics} {Benchmark} ({TPBench}): a
  dataset and study of {AI} reasoning capabilities in theoretical physics}.
\newblock \bibinfo{howpublished}{Preprint at
  \url{https://arxiv.org/abs/2502.15815}} (\bibinfo{year}{2025}).

\bibitem{mancoridis2025potemkin}
\bibinfo{author}{Mancoridis, M.}, \bibinfo{author}{Weeks, B.},
  \bibinfo{author}{Vafa, K.} \& \bibinfo{author}{Mullainathan, S.}
\newblock \bibinfo{title}{Potemkin understanding in large language models}.
\newblock \bibinfo{howpublished}{Preprint at
  \url{https://arxiv.org/abs/2506.21521}} (\bibinfo{year}{2025}).

\bibitem{deng2023k2}
\bibinfo{author}{Deng, C.} \emph{et~al.}
\newblock \bibinfo{title}{K2: a foundation language model for geoscience
  knowledge understanding and utilization}.
\newblock \bibinfo{howpublished}{Preprint at
  \url{https://arxiv.org/abs/2306.05064}} (\bibinfo{year}{2023}).

\bibitem{lin2024geogalactica}
\bibinfo{author}{Lin, Z.} \emph{et~al.}
\newblock \bibinfo{title}{{GeoGalactica}: a scientific large language model in
  geoscience}.
\newblock \bibinfo{howpublished}{Preprint at
  \url{https://arxiv.org/abs/2401.00434}} (\bibinfo{year}{2024}).

\bibitem{bi2024oceangpt}
\bibinfo{author}{Bi, Z.} \emph{et~al.}
\newblock \bibinfo{title}{{OceanGPT}: a large language model for ocean science
  tasks}.
\newblock In \emph{\bibinfo{booktitle}{Proceedings of the 62nd Annual Meeting
  of the Association for Computational Linguistics}} (\bibinfo{year}{2024}).
\newblock \bibinfo{note}{{arXiv}:2310.02031}.

\bibitem{thulke2024climategpt}
\bibinfo{author}{Thulke, D.} \emph{et~al.}
\newblock \bibinfo{title}{{ClimateGPT}: towards {AI} synthesizing
  interdisciplinary research on climate change}.
\newblock \bibinfo{howpublished}{Preprint at
  \url{https://arxiv.org/abs/2401.09646}} (\bibinfo{year}{2024}).

\bibitem{ovadia2024finetuning}
\bibinfo{author}{Ovadia, O.}, \bibinfo{author}{Brief, M.},
  \bibinfo{author}{Mishaeli, M.} \& \bibinfo{author}{Elisha, O.}
\newblock \bibinfo{title}{Fine-tuning or retrieval? {Comparing} knowledge
  injection in {LLMs}}.
\newblock In \emph{\bibinfo{booktitle}{Proceedings of the 2024 Conference on
  Empirical Methods in Natural Language Processing}}, \bibinfo{pages}{237--250}
  (\bibinfo{year}{2024}).
\newblock \bibinfo{note}{{arXiv}:2312.05934}.

\bibitem{allenzhu2023knowledge}
\bibinfo{author}{Allen-Zhu, Z.} \& \bibinfo{author}{Li, Y.}
\newblock \bibinfo{title}{Physics of language models: part 3.1, knowledge
  storage and extraction}.
\newblock \bibinfo{howpublished}{Preprint at
  \url{https://arxiv.org/abs/2309.14316}} (\bibinfo{year}{2023}).

\bibitem{lampinen2025generalization}
\bibinfo{author}{Lampinen, A.~K.} \emph{et~al.}
\newblock \bibinfo{title}{On the generalization of language models from
  in-context learning and finetuning: a controlled study}.
\newblock \bibinfo{howpublished}{Preprint at
  \url{https://arxiv.org/abs/2505.00661}} (\bibinfo{year}{2025}).

\bibitem{su2025rewardbridge}
\bibinfo{author}{Su, Y.} \emph{et~al.}
\newblock \bibinfo{title}{Crossing the reward bridge: expanding {RL} with
  verifiable rewards across diverse domains}.
\newblock \bibinfo{howpublished}{Preprint at
  \url{https://arxiv.org/abs/2503.23829}} (\bibinfo{year}{2025}).

\bibitem{deepseek2025}
\bibinfo{author}{Guo, D.} \emph{et~al.}
\newblock \bibinfo{title}{{DeepSeek-R1} incentivizes reasoning in {LLMs}
  through reinforcement learning}.
\newblock \emph{\bibinfo{journal}{Nature}} \textbf{\bibinfo{volume}{645}},
  \bibinfo{pages}{633--638} (\bibinfo{year}{2025}).

\bibitem{liu2025prorl}
\bibinfo{author}{Liu, M.} \emph{et~al.}
\newblock \bibinfo{title}{{ProRL}: prolonged reinforcement learning expands
  reasoning boundaries in large language models}.
\newblock \bibinfo{howpublished}{Preprint at
  \url{https://arxiv.org/abs/2505.24864}} (\bibinfo{year}{2025}).

\bibitem{yue2025}
\bibinfo{author}{Yue, Y.} \emph{et~al.}
\newblock \bibinfo{title}{Does reinforcement learning really incentivize
  reasoning capacity in {LLMs} beyond the base model?}
\newblock In \emph{\bibinfo{booktitle}{Advances in Neural Information
  Processing Systems}} (\bibinfo{year}{2025}).
\newblock \bibinfo{note}{{arXiv}:2504.13837}.

\bibitem{wu2025leash}
\bibinfo{author}{Wu, F.} \emph{et~al.}
\newblock \bibinfo{title}{The invisible leash: why {RLVR} may or may not escape
  its origin}.
\newblock \bibinfo{howpublished}{Preprint at
  \url{https://arxiv.org/abs/2507.14843}} (\bibinfo{year}{2025}).

\bibitem{matsutani2025squeezes}
\bibinfo{author}{Matsutani, K.} \emph{et~al.}
\newblock \bibinfo{title}{{RL} squeezes, {SFT} expands: a comparative study of
  reasoning {LLMs}}.
\newblock In \emph{\bibinfo{booktitle}{International Conference on Learning
  Representations}} (\bibinfo{year}{2026}).
\newblock \bibinfo{note}{{arXiv}:2509.21128}.

\bibitem{yan2025luffy}
\bibinfo{author}{Yan, J.} \emph{et~al.}
\newblock \bibinfo{title}{Learning to reason under off-policy guidance}.
\newblock \bibinfo{howpublished}{Preprint at
  \url{https://arxiv.org/abs/2504.14945}} (\bibinfo{year}{2025}).

\bibitem{li2025questa}
\bibinfo{author}{Li, J.} \emph{et~al.}
\newblock \bibinfo{title}{{QuestA}: expanding reasoning capacity in {LLMs} via
  question augmentation}.
\newblock In \emph{\bibinfo{booktitle}{International Conference on Learning
  Representations}} (\bibinfo{year}{2026}).
\newblock \bibinfo{note}{{arXiv}:2507.13266}.

\bibitem{zhang2025scaf}
\bibinfo{author}{Zhang, X.} \emph{et~al.}
\newblock \bibinfo{title}{{Scaf-GRPO}: scaffolded group relative policy
  optimization for enhancing {LLM} reasoning}.
\newblock In \emph{\bibinfo{booktitle}{International Conference on Learning
  Representations}} (\bibinfo{year}{2026}).
\newblock \bibinfo{note}{{arXiv}:2510.19807}.

\bibitem{nath2025guide}
\bibinfo{author}{Nath, V.} \emph{et~al.}
\newblock \bibinfo{title}{Adaptive guidance accelerates reinforcement learning
  of reasoning models}.
\newblock In \emph{\bibinfo{booktitle}{Advances in Neural Information
  Processing Systems}} (\bibinfo{year}{2025}).
\newblock \bibinfo{note}{{arXiv}:2506.13923}.

\bibitem{yan2026fest}
\bibinfo{author}{Yan, K.}, \bibinfo{author}{Schwing, A.~G.} \&
  \bibinfo{author}{Wang, Y.-X.}
\newblock \bibinfo{title}{Boosting reinforcement learning with verifiable
  rewards via randomly selected few-shot guidance}.
\newblock \bibinfo{howpublished}{Preprint at
  \url{https://arxiv.org/abs/2605.15012}} (\bibinfo{year}{2026}).

\bibitem{yu2025dapo}
\bibinfo{author}{Yu, Q.} \emph{et~al.}
\newblock \bibinfo{title}{{DAPO}: an open-source {LLM} reinforcement learning
  system at scale}.
\newblock \bibinfo{howpublished}{Preprint at
  \url{https://arxiv.org/abs/2503.14476}} (\bibinfo{year}{2025}).
\newblock \bibinfo{note}{NeurIPS 2025}.

\bibitem{chu2025sftrl}
\bibinfo{author}{Chu, T.} \emph{et~al.}
\newblock \bibinfo{title}{{SFT} memorizes, {RL} generalizes: a comparative
  study of foundation model post-training}.
\newblock In \emph{\bibinfo{booktitle}{Proceedings of the 42nd International
  Conference on Machine Learning}} (\bibinfo{year}{2025}).
\newblock \bibinfo{note}{Preprint at \url{https://arxiv.org/abs/2501.17161}}.

\bibitem{shao2025spurious}
\bibinfo{author}{Shao, R.} \emph{et~al.}
\newblock \bibinfo{title}{Spurious rewards: rethinking training signals in
  {RLVR}}.
\newblock \bibinfo{howpublished}{Preprint at
  \url{https://arxiv.org/abs/2506.10947}} (\bibinfo{year}{2025}).

\bibitem{lin2025ehrmind}
\bibinfo{author}{Lin, J.}, \bibinfo{author}{Wu, Z.} \& \bibinfo{author}{Sun,
  J.}
\newblock \bibinfo{title}{Training {LLMs} for {EHR}-based reasoning tasks via
  reinforcement learning}.
\newblock \bibinfo{howpublished}{Preprint at
  \url{https://arxiv.org/abs/2505.24105}} (\bibinfo{year}{2025}).

\bibitem{zhang2025interplay}
\bibinfo{author}{Zhang, C.}, \bibinfo{author}{Neubig, G.} \&
  \bibinfo{author}{Yue, X.}
\newblock \bibinfo{title}{On the interplay of pre-training, mid-training, and
  {RL} on reasoning language models}.
\newblock \bibinfo{howpublished}{Preprint at
  \url{https://arxiv.org/abs/2512.07783}} (\bibinfo{year}{2025}).

\bibitem{zelikman2022star}
\bibinfo{author}{Zelikman, E.}, \bibinfo{author}{Wu, Y.}, \bibinfo{author}{Mu,
  J.} \& \bibinfo{author}{Goodman, N.~D.}
\newblock \bibinfo{title}{{STaR}: bootstrapping reasoning with reasoning}.
\newblock In \emph{\bibinfo{booktitle}{Advances in Neural Information
  Processing Systems}} (\bibinfo{year}{2022}).
\newblock \bibinfo{note}{{arXiv}:2203.14465}.

\bibitem{ma2026relift}
\bibinfo{author}{Ma, L.} \emph{et~al.}
\newblock \bibinfo{title}{Learning what reinforcement learning can't:
  interleaved online fine-tuning for hardest questions}.
\newblock In \emph{\bibinfo{booktitle}{International Conference on Learning
  Representations}} (\bibinfo{year}{2026}).
\newblock \bibinfo{note}{Preprint at \url{https://arxiv.org/abs/2506.07527}}.

\bibitem{shrestha2026privileged}
\bibinfo{author}{Shrestha, S.} \& \bibinfo{author}{Tessier, A.}
\newblock \bibinfo{title}{Rethinking privileged information in on-policy
  self-distillation}.
\newblock \bibinfo{howpublished}{Preprint at
  \url{https://arxiv.org/abs/2608.18271}} (\bibinfo{year}{2026}).

\bibitem{li2025structure}
\bibinfo{author}{Li, D.} \emph{et~al.}
\newblock \bibinfo{title}{{LLMs} can easily learn to reason from
  demonstrations: structure, not content, is what matters!}
\newblock \bibinfo{howpublished}{Preprint at
  \url{https://arxiv.org/abs/2502.07374}} (\bibinfo{year}{2025}).

\bibitem{ruis2024procedural}
\bibinfo{author}{Ruis, L.} \emph{et~al.}
\newblock \bibinfo{title}{Procedural knowledge in pretraining drives reasoning
  in large language models}.
\newblock In \emph{\bibinfo{booktitle}{International Conference on Learning
  Representations}} (\bibinfo{year}{2025}).
\newblock \bibinfo{note}{{arXiv}:2411.12580}.

\bibitem{kalai2025}
\bibinfo{author}{Kalai, A.~T.}, \bibinfo{author}{Nachum, O.},
  \bibinfo{author}{Vempala, S.~S.} \& \bibinfo{author}{Zhang, E.}
\newblock \bibinfo{title}{Why language models hallucinate}.
\newblock \bibinfo{howpublished}{Preprint at
  \url{https://arxiv.org/abs/2509.04664}} (\bibinfo{year}{2025}).

\bibitem{song2025tax}
\bibinfo{author}{Song, L.}, \bibinfo{author}{Shi, T.} \& \bibinfo{author}{Zhao,
  J.}
\newblock \bibinfo{title}{The hallucination tax of reinforcement finetuning}.
\newblock In \emph{\bibinfo{booktitle}{Findings of the Association for
  Computational Linguistics: EMNLP 2025}} (\bibinfo{year}{2025}).
\newblock \bibinfo{note}{{arXiv}:2505.13988}.

\bibitem{kirichenko2025abstention}
\bibinfo{author}{Kirichenko, P.}, \bibinfo{author}{Ibrahim, M.},
  \bibinfo{author}{Chaudhuri, K.} \& \bibinfo{author}{Bell, S.~J.}
\newblock \bibinfo{title}{{AbstentionBench}: reasoning {LLMs} fail on
  unanswerable questions}.
\newblock \bibinfo{howpublished}{Preprint at
  \url{https://arxiv.org/abs/2506.09038}} (\bibinfo{year}{2025}).

\bibitem{damani2025rlcr}
\bibinfo{author}{Damani, M.} \emph{et~al.}
\newblock \bibinfo{title}{Beyond binary rewards: training {LMs} to reason about
  their uncertainty}.
\newblock In \emph{\bibinfo{booktitle}{International Conference on Learning
  Representations}} (\bibinfo{year}{2026}).
\newblock \bibinfo{note}{{arXiv}:2507.16806}.

\bibitem{agarwal2024gkd}
\bibinfo{author}{Agarwal, R.} \emph{et~al.}
\newblock \bibinfo{title}{On-policy distillation of language models: learning
  from self-generated mistakes}.
\newblock In \emph{\bibinfo{booktitle}{International Conference on Learning
  Representations}} (\bibinfo{year}{2024}).
\newblock \bibinfo{note}{Preprint at \url{https://arxiv.org/abs/2306.13649}}.

\end{thebibliography}

\section*{Acknowledgements}
This work was supported by the Guangdong Natural Science Foundation (grant no. 2025A1515012950) and the Hebei Major Science and Technology Support Plan (Environmental Management Special Grant, no. 252S3701D).

\section*{Author contributions}
M.R. designed the templates and the training experiments, developed the software and ran the experiments. Y.Z. ran experiments, compiled the formulas and their validity limits for the mechanisms, and contributed to writing the manuscript. C.M. conceptualised the study, conducted the literature review, created the visualisations, and took primary responsibility for writing and revising the manuscript. J.S. developed the methodology, supervised the research, and provided strategic oversight. All authors reviewed and approved the final manuscript.

\section*{Competing interests}
The authors declare no competing interests.

\clearpage
\includepdf[pages=-]{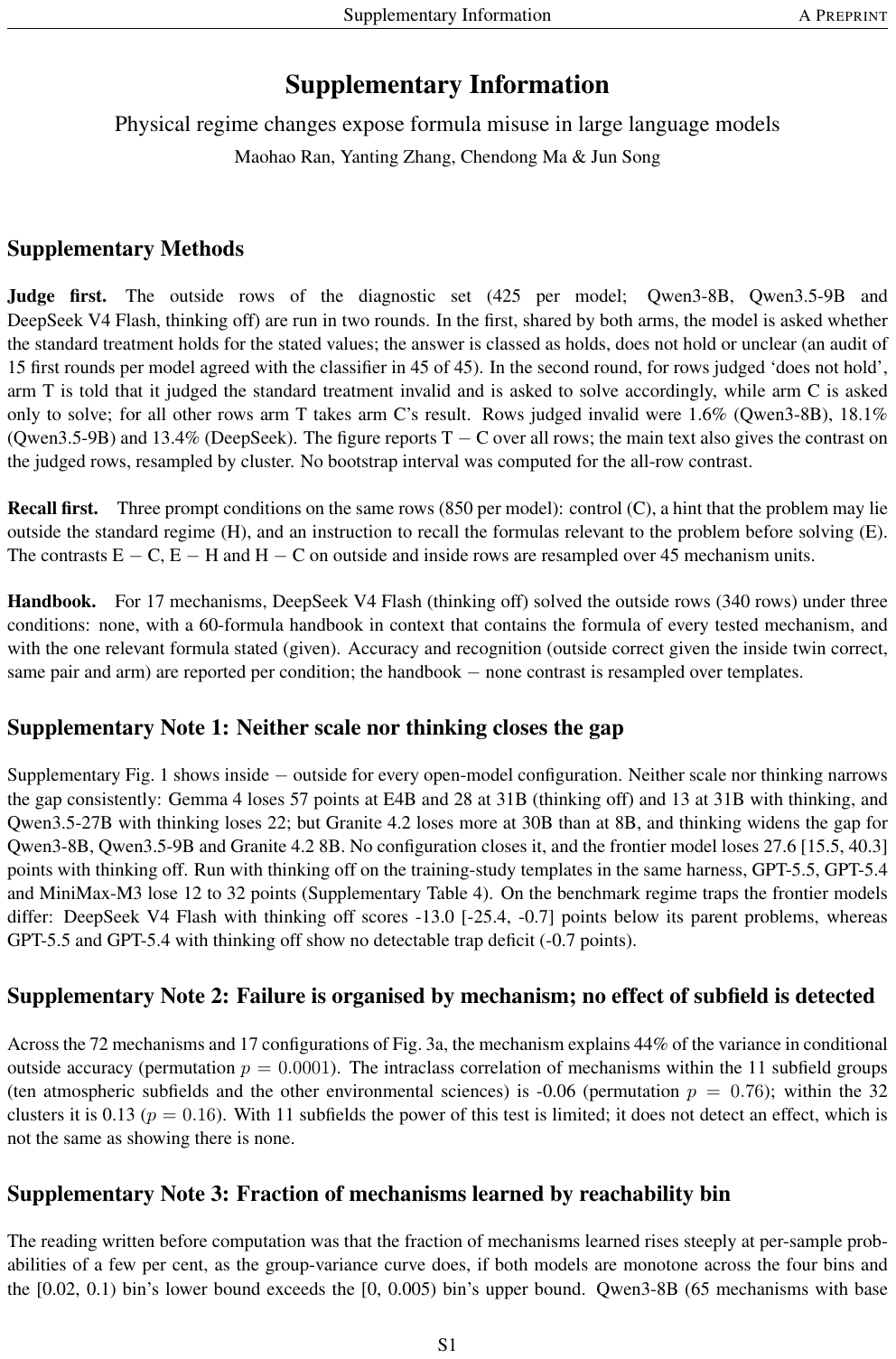}

\end{document}